%% file: main.tex
\documentclass{article}
\input{preamble}

\input{defs}
\begin{document}
\maketitle
\input{sec/0_abstract}

\input{sec/1_intro_tolga}

\input{sec/2_related_works_alp.tex}

\input{sec/2_background}
\input{sec/3_method_tolga}

\input{sec/4_experiments}

\input{sec/5_conclusion}

\bibliographystyle{plain}
\bibliography{main}

\appendix
\input{sec/6_appendix}

\newpage
\input{checklist}

\end{document}

%% file: preamble.tex
\usepackage[main, final]{neurips_2026}

\usepackage[utf8]{inputenc}
\usepackage[T1]{fontenc}
\usepackage{hyperref}
\usepackage{url}
\usepackage{booktabs}
\usepackage{amsfonts}
\usepackage{amsmath,amssymb,mathtools}
\usepackage{nicefrac}
\usepackage{microtype}
\usepackage{xcolor}
\usepackage{graphicx}
\usepackage{algorithm,algpseudocode}
\usepackage{booktabs}
\usepackage{multirow}
\usepackage{pifont}
\usepackage{bm}
\usepackage{mathrsfs}
\usepackage{wrapfig}
\usepackage{multirow}
\usepackage{float}
\usepackage{pifont}
\newcommand{\xmark}{\ding{55}}
\usepackage{booktabs, subcaption}
\usepackage[colorinlistoftodos,
            prependcaption,
            textsize=scriptsize]{todonotes}
\usepackage{enumitem}
\usepackage{cleveref}

\newcommand{\methodname}{HaNDF}

\title{HaNDF: Object-Conditioned\\ Geometric Neural Hand Distance Fields}

\author{%
  Zuhao Liu \quad
  Melih Darcan \quad
  Zhengdi Yu \quad
  Riza Alp Guler \quad
  Tolga Birdal \\
  Imperial College London
}

%% file: defs.tex
\usepackage{amsthm}

\renewcommand{\vec}[1]{\boldsymbol{#1}}

\newcommand{\x}{\mathbf{x}}

\newcommand{\R}{\mathbb{R}}

\newcommand{\Id}{\mathbf{I}}

\newcommand{\Rot}{\mathbf{R}}

\newcommand{\z}{\mathbf{z}}

\newcommand{\q}{\mathbf{q}}

\newcommand{\tb}{\mathbf{t}}

\newcommand{\ub}{\mathbf{u}}
\newcommand{\vb}{\mathbf{v}}

\newcommand{\pose}[0]{\vec{\theta}}
\newcommand{\shape}[0]{\vec{\beta}}

\newcommand{\Man}{\mathcal{M}}

\newcommand{\grad}[1]{\mathrm{grad}#1}
\newcommand{\SO}{\mathrm{SO}(3)}

\newcommand{\Quat}{\mathbb{H}}
\newcommand{\Quats}{\mathbb{H}_1}

\newcommand{\Exp}{\mathrm{{Exp}}}
\newcommand{\Log}{\mathrm{{Log}}}

\newcommand{\objects}{\mathcal{O}}
\newcommand{\obj}{O}
\newcommand{\latents}{\mathcal{Z}}

\newcommand\given[1][]{\:#1\vert\:}

\newcommand{\Amb}{\mathcal{X}}
\newcommand{\Skew}{\boldsymbol{\Omega}}
\DeclareMathOperator{\skewrm}{skew\!}

\DeclareMathOperator*{\argmin}{arg\min}

\newcommand{\eg}{\textit{e.g.}}

\newcommand{\etal}{\textit{et al.}}

\renewcommand{\paragraph}[1]{{\vspace{0.3mm}\noindent \bf #1}.}

\crefname{figure}{Fig.}{Figs.}
\Crefname{figure}{Fig.}{Figs.}
\crefname{table}{Tab.}{Tabs.}
\Crefname{table}{Tab.}{Tabs.}
\crefname{section}{Sec.}{Secs.}
\Crefname{section}{Sec.}{Secs.}
\crefname{equation}{Eq.}{Eqs.}
\Crefname{equation}{Eq.}{Eqs.}
\crefname{theorem}{Thm.}{Thms.}
\Crefname{theorem}{Thm.}{Thms.}
\crefname{remark}{Rem.}{Rems.}
\Crefname{remark}{Rem.}{Rems.}
\crefname{definition}{Dfn.}{Dfns.}
\Crefname{definition}{Dfn.}{Dfns.}
\crefname{algorithm}{Alg.}{Algs.}
\Crefname{algorithm}{Alg.}{Algs.}
\crefname{proposition}{Prop.}{Props.}
\Crefname{proposition}{Prop.}{Props.}

%% file: sec/0_abstract.tex
\begin{figure}[h]
  \centering
  \vspace{-7mm}
  \includegraphics[width=\linewidth]{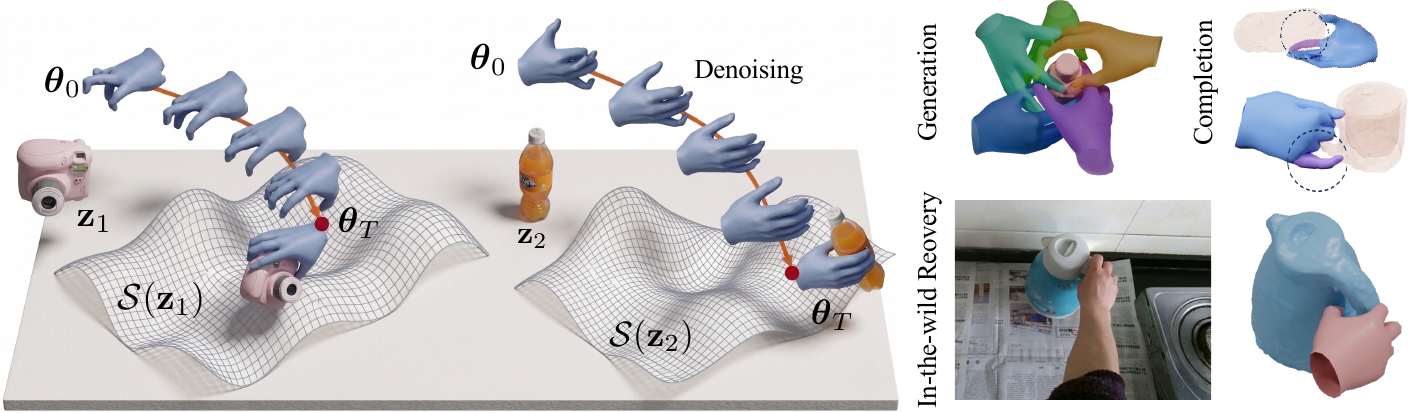}
  \caption{We introduce \textbf{HaNDF}, an object-conditioned generative hand prior. HaNDF is a geometric neural distance field representing the plausible hand configurations under interaction, and enables numerous tasks from denoising and completion to recovering hands or even objects from image data.\vspace{-3mm}}
  \label{fig:teaser}
\end{figure}

\begin{abstract}
We propose, \textbf{HaNDF}, object-conditional neural distance fields as data-driven priors for modeling the plausible hand–object configurations, in the product manifold of hand articulations. 
While recent unconditional generative priors showed success in modeling human pose, hands are versatile articulated bodies often found in interaction with other objects. This makes it crucial to develop a generative prior which can model the plausible hand poses conditioned on the object under interaction. To this end, our HaNDF proposes a geometry-aware, object-conditioned neural distance field, representing the plausible articulations in the zero-level set of a conditional field. We leverage tools from  Riemannian geometry to (i) \emph{represent} plausible hands in the zero level-set of an object-induced neural field, and (ii) \emph{project} a given pose onto this field. Using HaNDF, we can also optimize for a plausible \emph{condition}, determining the object under interaction. 
Our extensive evaluations demonstrate that our HaNDF provides a flexible and powerful prior for hand–object interaction, suitable for applications such as pose refinement, reconstruction under occlusion, and physically plausible manipulation synthesis.
\end{abstract}

%% file: sec/1_intro_tolga.tex
\vspace{-1.5mm}
\section{Introduction}
\label{sec:intro}
\vspace{-1.5mm}
Human hands are among the most versatile articulated systems, practicing rich and diverse interactions with objects. Modeling plausible hand–object configurations is therefore fundamental to applications in robotics, AR/VR, and 3D perception. Yet, recovering or generating such interactions from often partial observations remains challenging: the space of valid configurations is highly constrained, object-dependent, and non-Euclidean, where even small pose perturbations can lead to physically implausible grasps with interpenetration or loss of contact. A central requirement is thus a strong, queryable prior that encodes which hand poses are plausible given an object.

Existing approaches fall short of this goal. Generative models based on VAEs~\cite{rempe2021humor} or diffusion~\cite{shafir2023human,christen2024diffh2o} learn distributions over poses or interactions, but operate in Euclidean latent spaces where distances lack geometric meaning, limiting their expressivity. Neural distance field~\cite{chibane2020neural} (NDF) variants  such as PoseNDF~\cite{tiwari2022pose} or NRDF~\cite{he2024nrdf} offer an alternative by modeling the manifold of plausible articulations directly and enabling projection via gradient descent. However, current NDF-based priors are human-body specific, unconditional, capture hand articulation independent of the object, and thus fail to encode interaction-specific constraints. Conversely, object-conditioned methods~\cite{Kulkarni_2024_CVPR,weng2023neural} typically treat articulation in Euclidean space or rely on task-specific pipelines, sacrificing geometric consistency. As a result, no existing approach provides a unified prior that is (i) geometry-aware, (ii) conditioned on object shape, and (iii) equipped with a meaningful distance for optimization.

\textbf{HaNDF} addresses this gap proposing a generative, object conditioned prior of hand poses. Our key insight is to model hand articulation on its intrinsic Riemannian manifold while treating the object as a conditioning signal that modulates the manifold of plausible configurations. Concretely, we learn a conditional distance field that predicts the geodesic distance to the nearest valid hand pose given object geometry, whose zero level set defines the object-specific interaction manifold. This formulation yields a continuous, geometry-aware energy landscape where Riemannian gradients provide principled projection, refinement, and sampling directly on the articulation manifold.
This perspective unifies multiple applications under a single framework without retraining: denoising, recovering physically consistent grasps from data,  and generation of novel grasps. When integrated into reconstruction pipelines, HaNDF acts as a plug-and-play prior, improving robustness under occlusion without retraining task-specific models, as validated across three challenging benchmarks: HOI4D~\cite{liu2022hoi4d}, OakInk~\cite{yang2022oakink}, and DexYCB~\cite{chao2021dexycb}.
In summary, \textbf{our contributions are}:
\begin{itemize}[nosep,leftmargin=*]
    \item We introduce HaNDF, an object-conditioned Riemannian neural distance field for hand articulation, whose zero-level set defines a family of object-dependent plausibility manifolds.
    \item We develop an object-specific distance-field training procedure with per-object latent fitting and nearest-pose supervision, together with a transformer architecture that jointly processes hand-joint and object tokens.
    \item Based upon NRDF’s RDF-Grad, we enable object-conditioned projection, denoising, completion, and generation of MANO hand poses, and introduce an inversion procedure to retrieve and refine compatible object conditions.
\end{itemize}
HaNDF consistently outperforms state-of-the-art baselines, particularly in challenging settings with occlusion and ambiguity. Our code is publicly available at
\href{https://github.com/circle-group/HaNDF}{circle-group.github.io/research/HaNDF}.

%% file: sec/2_related_works_alp.tex
\vspace{-3mm}
\section{Related Work}
\label{sec:related}
\vspace{-3mm}
\paragraph{Generative priors of humans and hands}
Early efforts learned plausible-pose priors from 2D data with weak 3D supervision~\cite{omran2018neural}, used GANs~\cite{kanazawa2018end}, or extended adversarial supervision to motion sequences~\cite{kocabas2020vibe}.
Generative priors have been built on a range of architectures.
PoseD-Flow~\cite{nadar2026posedflow} trains a flow matching model of human pose.
MotionVAE~\cite{ling2020MVAE}, HuMoR~\cite{rempe2021humor}, and PhaseMP~\cite{shi2023phasemp} use VAEs autoregressively for motion priors.
Diffusion models~\cite{tevet2023human,shafir2023human,ci2023gfpose} have since become standard for pose and motion generation, with recent extensions to whole-body poses including hands~\cite{lu2025dposerx}.
Dyn-HaMR~\cite{yu2025dyn} trains an HMP prior for hand-motion infilling.
For hand-object interaction, TOCH~\cite{zhou2022toch} learns an object-centric correspondence-field prior for refining tracked sequences. 
Object-conditional grasp generators form a parallel
literature~\cite{jiang2021grasptta,ye2024g,cha2024text2hoi,zuo2024graspdiff,li2025latenthoi}.
These methods generate diverse poses but typically lack a differentiable distance to the pose manifold. We instead learn a geodesic distance to the conditioned plausible-grasp set, whose Riemannian gradient yields an projection direction. %

\paragraph{Neural distance fields (NDFs)}
NDFs were introduced for implicit representation of 3D objects and scenes~\cite{chibane2020neural} and have since become a tool for learning priors over articulated forms.
PoseNDF~\cite{tiwari2022pose} models body articulation as the zero-level set of an NDF.
NRDF~\cite{he2024nrdf} introduces a geometrically principled variant on the product of manifolds with better accuracy and diversity.
NRMF~\cite{yu2025nrmf} extends NRDF to temporal motion; Gu~\etal~\cite{gu2026hr} train NRDF on AMASS Retargeted-G1~\cite{mahmood2019amass} for humanoid retargeting.
We extend the conditional setting to human hands--objects, with a Riemannian distance field on the product manifold of joint quaternions and a learned object shape latent as condition.

\paragraph{Hand-object interaction reconstruction}
Hand-object reconstruction systems jointly recover hand and object geometry from monocular images.
Early parametric pipelines couple MANO with object pose regression~\cite{hasson2019learning}.
SDF-based methods replace meshes with implicit fields: Grasping Field~\cite{karunratanakul2020grasping}, AlignSDF~\cite{chen2022alignsdf}, gSDF~\cite{chen2023gsdf}, and HOISDF~\cite{qi2024hoisdf}.
Recent approaches leverage large pretrained models: diffusion priors for
object completion under
occlusion~\cite{liu2025easyhoi,fu2024d,wang2024Magichoi,aytekin2025follow,ye2024g}
and feed-forward transformers for direct point-cloud
prediction~\cite{chen2025hort}.
Our contribution is a hand prior, not a reconstruction pipeline: a differentiable distance to the conditioned plausible manifold whose gradient supports optimisation, denoising, completion, and refinement on top of any of the systems above.

%% file: sec/2_background.tex
\section{Background}
\label{sec:background}
Since hand joints are rotations, their natural configuration space is not Euclidean but a product of rotations. Here we focus on unit-quaternion representation of rotations and review the differential / Riemannian geometry of rotations and the product-manifold structure in~\cref{app:Riemannian}. 
Our exposition is based on NRDF~\cite{he2024nrdf} and NRMF~\cite{yu2025nrmf}, akin to~\cite{birdal2018bayesian,birdal2019probabilistic,chen2022projective,boumal2023introduction}.
In what follows, let $\Man$ be a smooth $m$-dimensional manifold embedded in an ambient Euclidean space $\Amb=\R^d$. For $\x\in\Man$, the \emph{tangent space} $T_{\x}\Man$ is the set of all velocity vectors of smooth curves passing through $\x$.

\paragraph{The rotation group as a Riemannian manifold}
The space of 3D rotations is a particular Riemannian manifold, known as the \emph{special orthogonal group}:
\begin{equation}
\SO
=
\left\{
\Rot\in\R^{3\times 3}
\;\middle|\;
\Rot^\top \Rot=\Id,\;\det(\Rot)=1
\right\}.
\end{equation}
Inspired by~\cite{he2024nrdf}, in this work, we parameterize rotations by \emph{unit quaternions} (UQ) and provide a detailed review of the group $\SO$ in~\cref{sec:app:rotations}.

A quaternion is an element $\q=(q_w,q_x,q_y,q_z)=(w,\vb)\in\R^4$, with scalar part $w\in\R$ and vector part $\vb\in\R^3$. The conjugate is $\bar{\q}=(w,-\vb)$, with the norm $\|\q\|_2=\sqrt{w^2+\|\vb\|_2^2}$.
The \emph{Hamilton product} of $\q=(w_{\q},\vb_{\q})$ and $\mathbf{r}=(w_{\mathbf{r}},\vb_{\mathbf{r}})$ is
\begin{equation}
\q\otimes \mathbf{r}
=
\left(
w_{\q}w_{\mathbf{r}}-\vb_{\q}^\top\vb_{\mathbf{r}},\;
w_{\q}\vb_{\mathbf{r}}+w_{\mathbf{r}}\vb_{\q}+\vb_{\q}\times\vb_{\mathbf{r}}
\right).
\end{equation}

\begin{wrapfigure}[12]{r}{0.3\linewidth}
\centering
\vspace{-4mm}
    \includegraphics[width=\linewidth]{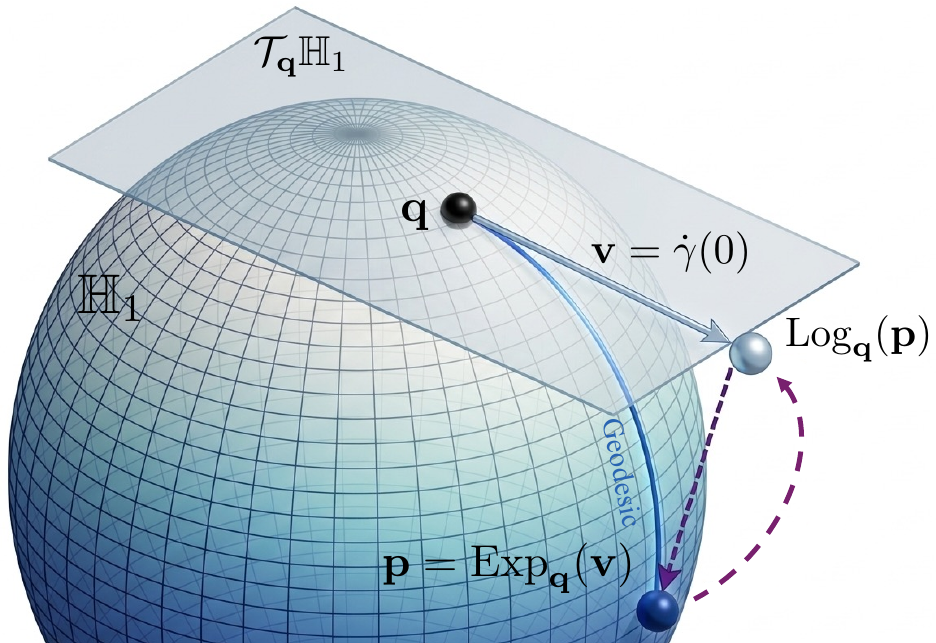}
    \caption{Manifold of quaternions.}
    \label{fig:quaternions}
\end{wrapfigure}

Shown in~\cref{fig:quaternions}, the unit quaternions form the three-sphere:
\begin{equation}
\Man_{\Quat}=\Quats
=
\left\{
\q\in\R^4
\;\middle|\;
\|\q\|_2=1
\right\}
\simeq S^3,
\end{equation}
and form a Lie group under Hamilton product. The inverse of a UQ is its conjugate: $\q^{-1}=\bar{\q}$. For a rotation by angle $\alpha$ around unit axis $\ub\in S^2$, the corresponding UQ is
\begin{equation}
\q=
\left(
\cos({\alpha}/{2}),
\;
\ub\sin({\alpha}/{2})
\right).
\end{equation}
The map from $\Quats$ to $\SO$ is a two-to-one (double) covering map: $\q \sim -\q$. That is, $\q$ and $-\q$ represent the same physical rotation. The tangent space at $\q$ is
\begin{equation}
T_{\q}\Quats
=
\left\{
\boldsymbol{\eta}\in\R^4
\;\middle|\;
\q^\top \boldsymbol{\eta}=0
\right\}.
\end{equation}
The orthogonal projection from ambient space $\R^4$ onto $T_{\q}\Quats$ is
\begin{equation}
\Pi_{\q}^{S^3}(\vb)
=
\vb-(\q^\top \vb)\q
=
(\Id-\q\q^\top)\vb.
\end{equation}

For a smooth $f:\Quats\to\R$, if $\nabla_{\q} f$ is the Euclidean gradient, the \emph{spherical Riemannian gradient} is
\begin{equation}
\grad_{\q} f
=
(\Id-\q\q^\top)\nabla_{\q} f.
\end{equation}

Let $\boldsymbol{\eta}\in T_{\q}\Quats$. In the quaternion Lie-group notation, tangent vectors are represented as pure imaginary quaternions $\boldsymbol{\eta}=(0,\ub)$ with $\ub\in\R^3$.
The \textbf{exponential} and \textbf{logarithmic maps} at $\q$ are:
\begin{align}
\Exp_{\q}(\boldsymbol{\eta})=\q\otimes \exp(\boldsymbol{\eta})=\q\otimes
\left(
\cos\vartheta,\;
\frac{\sin\vartheta}{\vartheta}\ub
\right),\quad
\Log_{\q}(\mathbf{p})=\left(0,\;\frac{\vb}{\|\vb\|_2}\cos^{-1}(w)
\right),
\end{align}
where $\boldsymbol{\eta}=(0,\ub)$ and $\vartheta=\|\ub\|_2$, and the \emph{group exponential at the identity} is
\begin{equation}
\exp(0,\ub)
=
\left(
\cos\|\ub\|_2,\;
\frac{\sin\|\ub\|_2}{\|\ub\|_2}\ub
\right),
\end{equation}
with the convention ${\sin\|\ub\|_2}/{\|\ub\|_2}=1$ when $\|\ub\|_2=0$ and $\Log_{\q}$ is zero when $\|\vb\|_2=0$, as used in NRDF~\cite{he2024nrdf}. 
The \textbf{Riemannian} (\textbf{geodesic}) {distance} between two unit quaternions $\q_1,\q_2\in\Quats$ is the length of the shortest geodesic on $S^3$, accounting for the antipodal identification $\q\sim -\q$:
\begin{align}
d_{\Quats}(\q_1,\q_2)=2\cos^{-1}\left(|\q_1^\top \q_2|\right)
= 2\min
\left\{
\|\Log_{\q_1}(\q_2)\|_2,\;
\|\Log_{\q_1}(-\q_2)\|_2
\right\}.
\end{align}

%% file: sec/3_method_tolga.tex
\section{\methodname: Geometric Object Conditioned Neural Hand Distance Fields}
\label{sec:method}

\paragraph{Hand representation}
We instantiate the articulated hand with the MANO model~\cite{mano}, which represents a hand mesh as a differentiable function of shape, pose, and global rigid motion:
\begin{equation}
\label{eq:MANO}
\mathbf{V}=\mathcal{H}(\shape, \pose, \tb)
\in \R^{N_v\times 3},
\end{equation}
where $\shape\in\R^{N_\beta}$ denotes low-dimensional hand shape, $\tb\in\R^3$ is global translation, and $\pose$ denotes the 3D articulated hand pose. In MANO, the pose consists of local rotations along a kinematic tree of hand joints. We exclude the global wrist rotation from the intrinsic articulation prior and parameterize the remaining $K=15$ local joint rotations on the power manifold of $\pose\vcentcolon=\{\q_i\in\Quats\}_{i=1}^K\in \Man_H=\Quats^K$. 
Each $\q_i$ parameterizes the rotation of joint $i$ with respect to its parent in the MANO kinematic tree. This \emph{relative}, \emph{local} parameterization is important in coupling the configuration of all local rotations.

\paragraph{Object representation}
Let $\mathcal{O}$ denote the space of 3D objects, represented as meshes, point clouds, or implicit fields. We assume access to an encoder $E : \objects \rightarrow \latents$, which maps an object $\obj\in \objects$ to a latent code: $\z = E(\obj) \in \latents$. Unlike the hand pose, $\mathcal{Z}\equiv\R^{d_z}$ is not endowed with a special Riemannian structure.
The encoder is trained jointly with a decoder $D : \latents \to \objects$ in an autoencoding framework. As we will detail in~\cref{sec:handf}, while any embedding network is possible, we adopt a DeepSDF-style~\cite{park2019deepsdf} architecture that produces a compact latent code capturing global object geometry.

We deliberately represent the object via a latent code rather than explicit geometry (\eg, point clouds or meshes) for reasons of:
(i) \emph{compactness}, enabling efficient conditioning in neural fields; (ii) \emph{invariance} to input resolution and representation; (iii) \emph{compatibility} with generative and reconstruction pipelines, where object shape may be partially observed.

\subsection{\methodname}
\label{sec:handf}
Given a hand pose $\pose \in \Man_H$ and an object latent $\z\in\latents$, our model learns a scalar distance field
\begin{equation}
f_\Phi : \Man_H \times \latents \to \R_{\ge 0},
\end{equation}
that explicitly approximates the geodesic distance to the nearest plausible hand pose conditioned on the object. The \emph{conditioned manifold} of plausible hands is thus given by the zero level-set:
\begin{equation}
\mathcal{S}(\z)
=
\left\{
\pose \in \Man_H \;\middle|\; f_\Phi(\pose\given\z) = 0
\right\},
\end{equation}
such that the value of $f$ represents the unsigned geodesic
distance to the closest plausible pose on the manifold.
The latent $\z$ influences the field globally, adapting the location and shape of the zero level-set $S(z)$
according to object geometry. Intuitively, $\z$ encodes affordances such as size, curvature, and topology, which determine feasible grasp configurations.

\noindent\textbf{Projection onto \methodname.}
Given an object $\z$, an arbitrary hand pose $\pose^{(0)}\in\Man_H$ can be projected onto the closest plausible hand pose on our object-conditioned manifold: 
\begin{equation}
\Pi_{\mathcal{S}(\z)}(\pose^{(0)})
=
\argmin_{\pose\in\mathcal{S}(\z)}
d_{\Man_H}(\pose,\pose^{(0)}).
\end{equation}
Since $\mathcal{S}(\z)$ is represented implicitly by the zero level-set of $f_\Phi$, we solve this optimization by minimizing the predicted distance field over the hand manifold, similar to~\cite{tiwari2022pose,he2024nrdf}:
\begin{equation}
\pose^\star \approx \argmin_{\pose\in\Man_H}
f_\Phi(\pose\given\z).
\end{equation}
Since $\z$ is a condition, the optimization is performed only over $\pose$. We then update the hand pose using an \emph{adaptive conditioned Riemannian descent step} (see~\cref{fig:teaser} for iterations):
\begin{equation}
\pose^{(t+1)}
=
\Exp_{\pose^{(t)}}
\left(
-\alpha
f_\Phi(\pose^{(t)}\given\z)
\frac{
\grad_{\pose} f_\Phi(\pose^{(t)}\given\z)
}{
\left\|
\grad_{\pose} f_\Phi(\pose^{(t)}\given\z)
\right\|_{\pose^{(t)}}+\epsilon
}
\right),\tag{HaNDFGrad}
\label{eq:HaNDFGrad}
\end{equation}
where $\alpha>0$ is a step-size parameter and $\epsilon>0$ prevents numerical instability. By construction, \cref{eq:HaNDFGrad} respects the non-Euclidean manifold of MANO hand articulations as every update remains on $\Man_H=\Quats^K$, and the multiplication by $f_\Phi(\pose^{(t)}\given\z)$ makes the step length proportional to the predicted distance, akin to the RDFGrad of NRDF~\cite{he2024nrdf}. The projection is terminated when either the predicted distance falls below a threshold $f_\Phi(\pose^{(t)}\given\mathbf{z}) < \delta$ or maximum iterations is reached.

\begin{wrapfigure}[15]{r}{0.2\linewidth}
\centering
    \includegraphics[width=\linewidth]{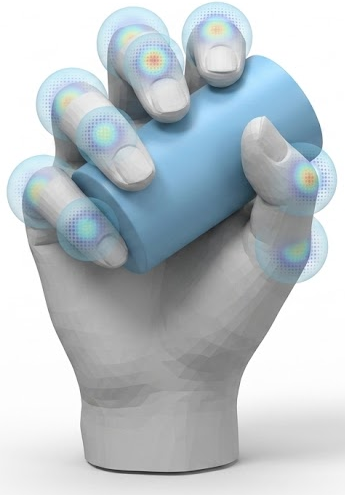}
    \caption{HaNDF jointly models the plausible configurations of joints given the object code.}
    \label{fig:distr}
\end{wrapfigure}
\noindent\textbf{\methodname~as a conditional generative model.} 
Beyond projection and refinement, \methodname~can be used as a conditioned generative model of plausible hand poses. For a fixed latent $\z$, the $f$-induced Bolzmann density $p(\pose\given\z) \propto \exp(-\beta f_\phi(\pose\given\z)^2)$, approximates data distribution,  understood as a product-measure over spheres, illustrated in~\cref{fig:distr}, analogous to~\cite{yu2025nrmf}. \cref{eq:HaNDFGrad} is then a maximum-aposteriori estimation. 

Our simple strategy to sample from $p(\cdot\given\z)$ is as follows: we (i) initialize a random hand pose $\pose^{(0)}$ by sampling each quaternion independently, $\q_k^{(0)} \sim \mathrm{Unif}(\Quats) \, \forall k$ or sampling $\tilde{\q}_k\sim\mathcal{N}(\mathbf{0},\mathbf{I}_4)$ and normalizing $\q_k^{(0)}={\tilde{\q}_k}/{\|\tilde{\q}_k\|_2}$, (ii) project it onto \methodname, and (iii) pass it from MANO model in~\cref{eq:MANO}.   
Different initial samples can project to different connected components or basins of attraction of $\mathcal{S}(\mathbf{z})$. Hence, generation diversity is controlled by the initial distribution and by the geometry of the learned field.

A connection to Riemannian Flow Matching~\cite{chenflow} has been made in~\cite{he2024nrdf} for the unconditional NRDF. Our distance field formulation has some advantages. First, it avoids imposing a Gaussian structure on the hand-poses. Instead of predicting the step towards the manifold via a network, we obtain it via backpropagation. This allows optimization in domains where designing decoders is challenging. Also, we can utilize existing optimizers of deep learning frameworks for Lagrangian iterations by simply minimizing distance.

\paragraph{\methodname~as a generative prior} 
The same learned field can be used for generation, denoising, completion, and reconstruction, since all tasks reduce to minimizing or projecting through the same object-conditioned distance field. Typically, in downstream tasks, our prior is combined with task-specific data terms involving observations $\mathcal{Y}$, leading to the inference problem\footnote{In the language of diffusion or flow matching, such inference is understood as \emph{inversion} or \emph{training-free guidance}~\cite{nadar2026posedflow,ben2024d}.}:
\begin{equation}
\min_{\pose\in\Man_H,\mathbf{z}\in\mathcal{Z}}
\quad
\mathcal{L}_{\mathrm{data}}(\pose,\mathbf{z}\given\mathcal{Y})
+
\lambda_{\mathrm{prior}}
f_\Phi(\pose\given\mathbf{z})^2
+
\lambda_z \mathcal{R}(\z),
\end{equation}
which can be minimized for $\pose\in\Man_H$ (Riemannian Adam~\cite{becigneul2018riemannian}) to recover hands or for $\z$ (Euclidean) to recover objects that best fit the hand. 
The regularizer $\mathcal{R}(\z)$ prevents degenerate latent drift and may be given by a reconstruction loss through the object decoder. $\mathcal{L}_{\mathrm{data}}$ is task-dependent and can ensure consistency with observed depth, silhouette, 3D measurements. 

\paragraph{Object encoding}
We represent the object as a single global shape code $\mathbf{z}\in\mathbb{R}^{c_o}$ produced by a pretrained DeepSDF auto-decoder~\cite{park2019deepsdf}. Following DeepSDF, each shape is paired with an continuous latent code $\mathbf{z}$, and the decoder $g_\psi(\mathbf{z}, \mathbf{x})$ regresses the signed distance at any query point. We choose DeepSDF auto-decoder for two reasons: (i) its latent space is inherently aligned with our object condition formulation, and (ii) its decoder-only design supports geometry refinement via latent optimization at test time. Note that our object guidance only requires a permutation-invariant representation of the object and is therefore compatible with alternative encoders~\cite{cho2025representing,yang2025pandora3d}. 

To recover objects at inference, we invert (auto-decode) to obtain the DeepSDF latent code $K$ times, starting from different initializations and retrieve the nearest training latents, and keep hand-object pair with smallest error. We use PCA to reduce the dimensionality of this search. We provide the details and comparisons to other encoders in~\cref{app:object_retrieval}.

\paragraph{\methodname~architecture}
We tokenize hand and object into a shared $d$-dimensional space and process them with a transformer encoder. The $K$ joint quaternions are linearly projected to hand tokens $\mathbf{h}_k=\mathbf{W}_h\mathbf{q}_k+\mathbf{b}_h$. We project the DeepSDF latent code to obtain one object token $\mathbf{o}=\mathbf{W}_o\mathbf{z}+\mathbf{b}_o$. A learnable \texttt{[DIST]} token $\mathbf{d}$ is prepended as an aggregator, yielding the input sequence
\begin{equation}
\mathbf{X} \;=\; [\,\mathbf{d};\; \mathbf{h}_1,\ldots,\mathbf{h}_{K};\; \mathbf{o}\,]\;\in\;\mathbb{R}^{(K+2)\times d}.
\end{equation}
After learnable positional embeddings, $\mathbf{X}$ is processed by $L$ self-attention layers; the final-layer representation at the \texttt{[DIST]} position is mapped to the scalar prediction by a linear head with softplus function,
\begin{equation}
f_\Phi(\boldsymbol{\theta}\given\mathbf{z}) \;=\; \mathrm{softplus}(\mathbf{w}^\top\mathbf{d}^{(L)} + b).
\end{equation}
Self-attention lets every hand-joint token attend to the object token and vice versa, so the prediction is jointly conditioned on hand articulation and object shape. %
Encoding the entire shape code as a single token preserves its holistic semantics and lets each joint query the full object descriptor in one attention hop; we use $L=4$ layers and $c_o=64$, with both choices justified in~\cref{tab:ablation_arch}. MLP is also an alternative to build the architecture, as shown in~\cref{app:arch}.

\vspace{-1mm}
\paragraph{Training}
We train $f_\Phi$ using the regression loss on paired hand--object data with synthetic noise applied to the hand pose. The training distribution must place sufficient mass on near-manifold (hard-negative) samples since at inference our projection spends most of its iterations around the \emph{narrow band}. Hence, we inherit the exponential sampling schedule from~\cite{he2024nrdf} for hands. The exponential distribution assigns higher density to small magnitudes than to large ones, with non-zero density throughout the tail. We also sample the conditioning object over the epochs proportional to the plausible grasps per object provided in the dataset. We provide the full training algorithm in~\cref{alg:handf_training}.

%% file: sec/4_experiments.tex
\vspace{-2mm}
\section{Experiments}
\label{sec:experiments}
\vspace{-2mm}

To evaluate the effectiveness of the proposed \methodname{}, we conduct experiments on generation, denoising, in-the-wild hand recovery, and hand--object recovery, using task-specific evaluation metrics. We evaluate our method on several datasets, including HOI4D~\cite{liu2022hoi4d}, OakInk~\cite{yang2022oakink}, and DexYCB~\cite{chao2021dexycb}. We also provide detailed ablation studies on the network architecture, object encoding, and field geometry, along with further analyses of runtime, robustness, and generalization.

\paragraph{Datasets}
We evaluate our method on three large-scale hand–object interaction datasets. \textbf{HOI4D} \cite{liu2022hoi4d} provides egocentric RGB-D sequences with dense MANO annotations, object poses, and masks. We follow an object-level split across six rigid categories, using 219 objects (63K poses) to train and 30 unseen objects (7.5K poses) to test. We retain only contact frames. \textbf{DexYCB} \cite{chao2021dexycb} contains 1,000 third-person grasp sequences over 21 YCB objects with full MANO and object pose annotations, from which we use the standard \texttt{s0} split (582K frames). \textbf{OakInk} \cite{yang2022oakink} complements these with diverse hand–object interactions in multi-view RGB sequences. We use sequences from 7 object categories to build our training and test sets, with 53 objects for training and 14 unseen objects for testing. More details in \cref{app:datasets}. %

\paragraph{Baselines}
We compare against three groups of baselines. First, we adapt pose distance-field priors to the hand domain by re-implementing PoseNDF~\cite{tiwari2022pose} and NRDF~\cite{he2024nrdf} on the 15-joint MANO manifold, denoted H-PoseNDF (Euclidean updates with quaternion normalization) and H-NRDF (Riemannian formulation on $(\mathbb{S}^3)^{15}$), both of which are unconditional and serve as direct ablations of our object conditioning. Second, we compare to recent hand and hand–object generative priors, including HMP~\cite{duran2024hmp} and object-conditioned models such as G-HOP~\cite{ye2024g}, LatentHOI~\cite{li2025latenthoi}, and Text2HOI~\cite{cha2024text2hoi}. Third, for single-image hand recovery, we evaluate against EasyHOI~\cite{liu2025easyhoi}, AlignSDF~\cite{chen2022alignsdf}, and IHoI~\cite{ye2022s}, providing ground-truth object geometry to all methods to isolate hand reconstruction, and integrating our method as an object-conditioned refinement prior within EasyHOI and AlignSDF.
\cref{app:baselines} provides further information.

\paragraph{Metrics}
We adopt a unified set of evaluation metrics across all downstream tasks,
covering four complementary aspects: realism,
distribution fidelity, sample diversity, and
pose reconstruction accuracy. All angular metrics are computed
on the product manifold $(\mathbb{S}^3)^{15}$ formed by the 15 MANO
finger-joint quaternions and reported in degrees; Euclidean metrics
are reported in millimeters (mm) or centimeters (cm) as indicated, and F-scores are reported at distance thresholds expressed as percentages of the object's longest bounding-box half-side. A summary of our metrics are given below and~\cref{app:metrics} provides further details:
\begin{itemize}[nosep,leftmargin=*]
    \item \textbf{$\Delta_q$}: Geodesic rotation error with GT hand across all 15 joints  (degrees)
    \item \textbf{dNN}: Geodesic distance to nearest GT pose in train set under the same object catogory (degrees)
    \item \textbf{m2m}: Mean Euclidean distance between predicted and ground-truth 3D vertices (mm)
    \item \textbf{MPVPE / MPJPE}: Mean per-vertex / per-joint position error on MANO mesh (mm)
    \item \textbf{I.V.}: Intersection volume between hand mesh and object mesh (cm$^3$)
    \item \textbf{Chamfer}: Chamfer distance between the reconstructed and ground-truth objects in normalized units.
    \item \textbf{FID}: Fréchet Inception Distance for pose distribution matching
    \item \textbf{APD}: Average pairwise geodesic distance measuring generation diversity (degrees)
\end{itemize}
Generation uses \{\textit{FID}, \textit{APD}, \textit{dNN}\};
denoising \{$\Delta_q$, \textit{dNN}, \textit{m2m}\};
completion \{$\Delta_q$, \textit{dNN},
\textit{APD}, \textit{FID}\}; recovery
\{\textit{MPJPE}, \textit{MPVPE}, \textit{I.V.}\}, and
\{\textit{obj-CD}, \textit{obj-F@}$\tau$\} when object geometry is
optimized.

\vspace{-2mm}
\subsection{Results on Downstream Tasks}
\vspace{-2mm}

\begin{wrapfigure}[15]{r}{0.4\linewidth}
\centering
\vspace{-1mm}
   \includegraphics[width=\linewidth]{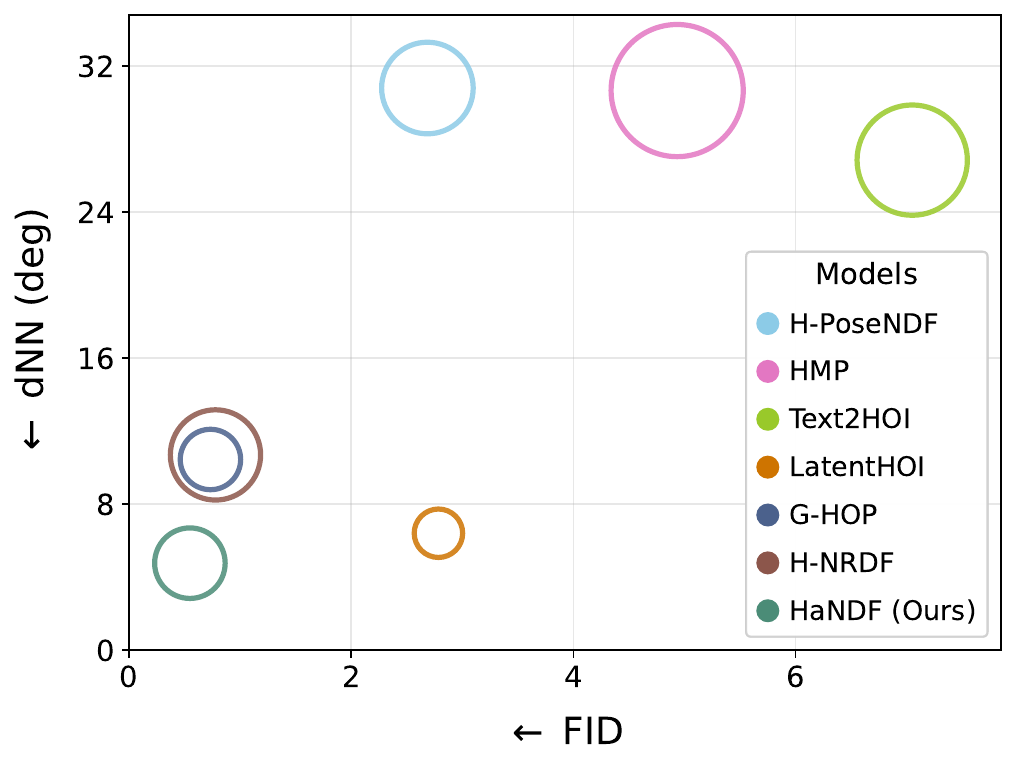}
        \vspace{-4mm} %
        \captionof{figure}{Pose generation on HOI4D: FID vs. dNN (lower is better); bubble size encodes APD.}
        \label{fig:generation}
\end{wrapfigure}

\textbf{Grasp pose generation.}
Given an object instance represented by its canonical geometry $\mathcal{O}$,
the goal is to generate hand poses $\boldsymbol{\theta}_h$ that plausibly
grasp $\mathcal{O}$. A successful generator should produce hand poses that
are (i) \emph{plausible} as natural human grasps,
(ii) \emph{compatible} with the conditioning object's geometry, and
(iii) \emph{diverse}, covering the many ways a human would naturally
grasp the same object.

As shown in \cref{fig:generation}, \methodname{} lands in
the bottom-left of the FID--dNN plane while maintaining a competitive
bubble size, dominating both unconditional priors (H-NRDF, H-PoseNDF) and
conditional generative baselines. As detailed in \cref{tab:generation}, our method achieves the best dNN (\textbf{4.77$^\circ$}) and a highly competitive FID (0.54), indicating that our generated poses are both realistic and well-aligned with the data distribution. Although H-NRDF also reports low FID (0.77), its much higher dNN reveals limited per-sample realism. Compared to LatentHOI, \methodname{} attains substantially higher diversity (APD 12.73$^\circ$ vs.\ 6.01$^\circ$) while also improving FID and dNN, demonstrating a better balance between fidelity and diversity. We provide more pose generation results in~\cref{app:hand_generation_app}.

\paragraph{Hand pose denoising}
We evaluate our hand pose denoising on the HOI4D test split with novel objects per category. For each clean test
pose, we add noise at different levels $\sigma$, producing
$n{=}3000$ noisy samples per $\sigma$; identical noisy inputs are
fed to all methods to ensure an fair comparison. We
report three metrics (lower-is-better for $\Delta_q$,
\textit{dNN}, and \textit{m2m}). All distance-field methods (\methodname, H-NRDF,
H-PoseNDF) denoise by gradient descent on the learned scalar field
for the same iterations with the same step size.  Except our method, none of the baselines can consider the object condition.
We provide further results with varying object encoders in~\cref{app:resultsObjects}, and more hand denosing results can be found in~\cref{sec:hand_denoising_app}.

\begin{figure}[t]
    \centering
    \includegraphics[width=\linewidth]{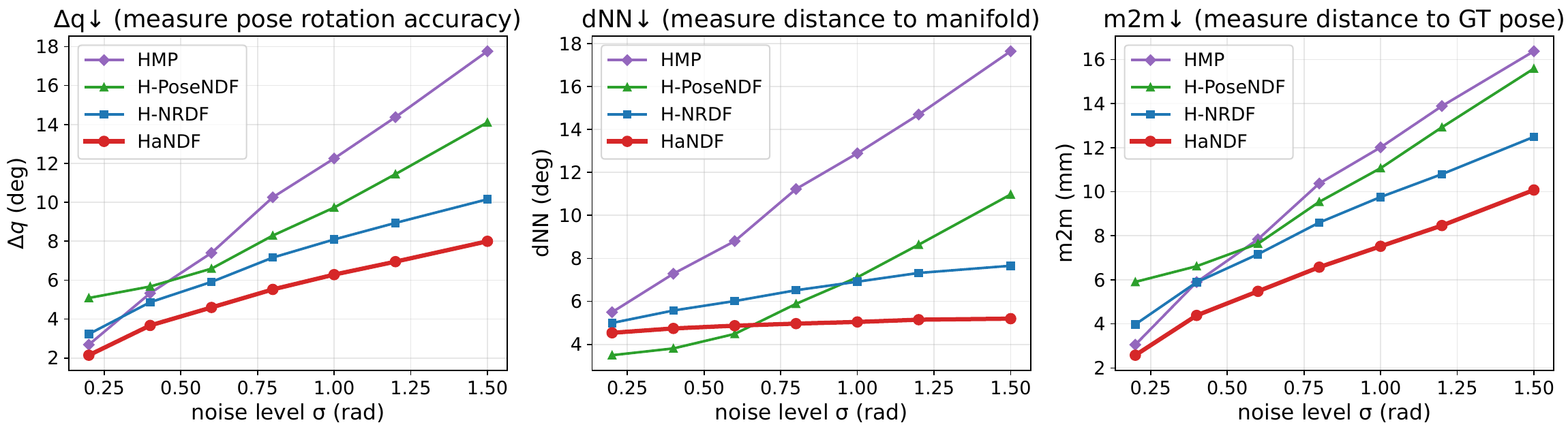}
    \caption{
        \textbf{Hand-pose denoising} on HOI4D under different noise levels ($\sigma\in$ [0.2, 1.5] rad). \methodname\ (ours, red) achieves the best $\Delta_q$ and m2m at every noise level, with margins growing under heavier corruption.\vspace{-5mm}}
    \label{fig:denoising}
\end{figure}

\begin{figure}[hb]
    \centering
    \includegraphics[width=\linewidth]{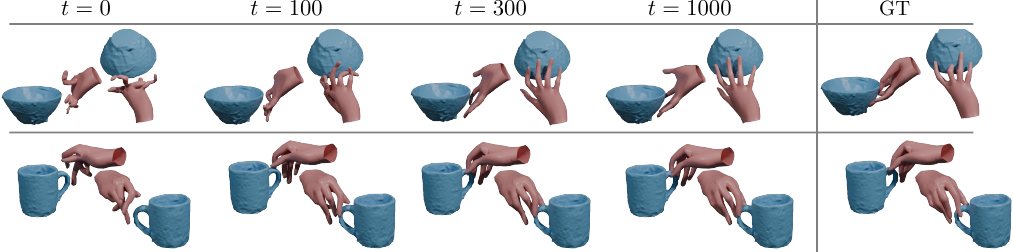}
    \caption{Qualitative pose denoising iterations for hand-object interaction on HOI4D. Starting from noisy initializations ($t=0$), our model iteratively refines the hand poses. By $t=1000$, the hand reaches a physically plausible configuration that closely matches the Ground Truth (GT).\vspace{-3mm}}
    \label{fig:qual-denoising-process}
\end{figure}

\begin{figure}[t]
    \centering
    \includegraphics[width=\linewidth]{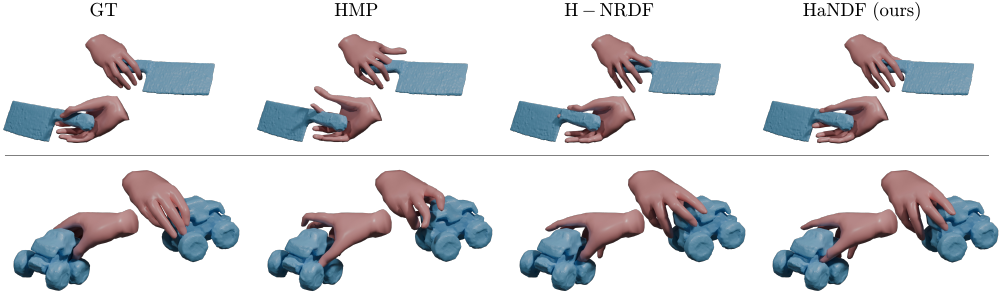}
    \caption{Qualitative comparison of pose denoising across different methods on HOI4D. We evaluate our method (\methodname) against HMP and H-NRDF on a variety of object geometries. Ground Truth (GT) is provided for reference.\vspace{-3mm}}
    \label{fig:qual-denoising}
\end{figure}

As shown in \cref{fig:denoising} (supplemented by \cref{tab:hand_denoising}),
\methodname\, achieves the best $\Delta_q$ and m2m at every
noise level, with margins that widen as $\sigma$ grows. Most
strikingly, \methodname's \textit{dNN} stays nearly constant across
the full noise range ($4.54^\circ \rightarrow 5.20^\circ$). 
On dNN, \methodname\ dominates from $\sigma$ $\geq$ 0.8 onward. At the smallest two $\sigma$, H-PoseNDF sits slightly closer to the manifold, but at the cost of much higher $\Delta_q$ and m2m. This is a mode-collapse signature, where outputs cluster near a few training centroids rather than tracking the input.
Qualitative comparisons in \cref{fig:qual-denoising-process} and \cref{fig:qual-denoising}  further validate these findings, showing that \methodname\ consistently recovers plausible poses.

\paragraph{Hand completion}
We now evaluate (object-conditioned) hand pose completion on the HOI4D test split under two masking protocols: \textbf{Missing Fingers}, where one of the 5 fingers
(3 joints) is randomly masked per sample with the other 12 joints
observed at GT; and \textbf{Missing Finger Parts}, where only the 5
finger-base joints are observed and the remaining 10 
joints are masked. The masked joints are initialized with the same
exponential noise, and $n{=}3000$ samples
are drawn per task with identical inputs fed to all methods.
Inference performs gradient descent on the learned scalar field, updating \emph{only} the masked joints and keeping
observed joints frozen. To isolate completion
quality from the trivially-correct observed joints. All metrics are
computed on the masked-joint subset including $\Delta_q$,
\textit{dNN}, \textit{APD} , and \textit{FID}.

\begin{table}[t]
\centering
  \caption{\textbf{Hand part completion} across two settings: missing fingers and partly missing fingers on HOI4D dataset.}
  \label{tab:comparison}
  \small
  \setlength{\tabcolsep}{8.5pt}
  \begin{tabular}{lcccc cccc}
    \toprule
    & \multicolumn{4}{c}{Missing Fingers} & \multicolumn{4}{c}{Missing Finger Parts} \\
    \cmidrule(lr){2-5} \cmidrule(lr){6-9}
    Method & $\Delta_q \downarrow$ & dNN$\downarrow$ & APD$\uparrow$ & FID$\downarrow$ & $\Delta_q\downarrow$ & dNN$\downarrow$ & APD$\uparrow$ & FID$\downarrow$ \\
    \midrule
    HMP \cite{duran2024hmp}                           & 24.37$^\circ$          & 18.77$^\circ$          & \textbf{36.86$^\circ$}          & 0.7407          & 16.60$^\circ$          & 14.45$^\circ$          & \textbf{26.79$^\circ$}          & 0.9448 \\
    H-PoseNDF            & 20.80$^\circ$          & 16.40$^\circ$          & 32.57$^\circ$          & 1.0318          & 12.87$^\circ$          & 8.83$^\circ$           & 20.03$^\circ$          & 0.7267 \\
    H-NRDF    & 16.60$^\circ$          & 8.01$^\circ$           & 24.48$^\circ$          & 0.1558          & 8.47$^\circ$           & 4.52$^\circ$           & 12.17$^\circ$ & 0.0453 \\
    \methodname       & \textbf{12.57$^\circ$} & \textbf{5.44$^\circ$}  & 20.79$^\circ$          & \textbf{0.0596} & \textbf{6.60$^\circ$}  & \textbf{3.57$^\circ$}  & 12.23$^\circ$          & \textbf{0.0106} \\
    \bottomrule
  \end{tabular}
  \vspace{-4mm}
\end{table}

As reported in \cref{tab:comparison}, \methodname\, achieves the
best $\Delta_q$, \textit{dNN}, and \textit{FID} on both completion
tasks, cutting $\Delta_q$ by $\sim$25\% and \textit{FID} by
3--4$\times$ over the strongest baseline (H-NRDF), with even larger
gaps against H-PoseNDF and HMP. The unconditional baselines follow a
consistent ranking H-NRDF $>$ H-PoseNDF $>$ HMP. Notably, H-NRDF's gap to \methodname\, widens on Missing Fingers (\textit{dNN} 8.01$^\circ$ vs.\ 5.44$^\circ$) compared to Missing Finger Parts (4.52$^\circ$ vs.\ 3.57$^\circ$), highlighting that \methodname's object conditioning compensates when more joints anchor the pose.

\begin{table}[h]
  \centering
  \caption{\textbf{In-the-wild hand recovery} on HOI4D, OakInk, and DexYCB. For hand-focused, fair comparisons, the object ground truth is provided during inference. Lower is better.}
  \label{tab:sota_comparison}
  \resizebox{\linewidth}{!}{%
  \begin{tabular}{l|ccc|ccc|ccc}
    \toprule
    & \multicolumn{3}{c|}{HOI4D~\cite{liu2022hoi4d}} & \multicolumn{3}{c|}{OakInk~\cite{yang2022oakink}} & \multicolumn{3}{c}{DexYCB~\cite{chao2021dexycb}} \\
    Method  & MPVPE$\downarrow$ & MPJPE$\downarrow$ & I.V.$\downarrow$
            & MPVPE$\downarrow$ & MPJPE$\downarrow$ & I.V.$\downarrow$
            & MPVPE$\downarrow$ & MPJPE$\downarrow$ & I.V.$\downarrow$ \\
    \midrule
    IHoI~\cite{ye2022s} & 19.42 & 20.12 & 41.44 & 14.75 & 15.17 & 1.66 & 29.85 & 30.73 & 1.26 \\
    EasyHOI~\cite{liu2025easyhoi} & 16.64 & 17.06 & 13.28 & 13.17 & 13.48 & 0.93 & 29.64 & 30.08 & 2.61 \\
    EasyHOI + \methodname{} & 13.98  & 14.33 & 11.40 & 12.85 & 13.12 & 0.89 & 28.31 & 28.78 & 2.43 \\
    AlignSDF~\cite{chen2022alignsdf} & 10.40 & 10.60 & 5.39 & 10.34 & 10.50 & \textbf{0.82} & 16.82 & 17.19 & 0.99 \\
    AlignSDF + \methodname{} & \textbf{10.22} & \textbf{10.43} & \textbf{4.39} & \textbf{9.98} & \textbf{10.17} & 0.87 & \textbf{16.59} & \textbf{16.97} & \textbf{0.98} \\
    \bottomrule
  \end{tabular}}
\end{table}

\paragraph{Hand-object reconstruction}
Single-image hand--object reconstruction remains a challenging task due to severe mutual occlusion, and serves as a natural downstream application of our prior. Since existing methods adopt heterogeneous object reconstruction pipelines (e.g.\ EasyHOI~\cite{liu2025easyhoi} relies on LISA + InstantMesh, while AlignSDF~\cite{chen2022alignsdf} uses a trained implicit SDF), we provide the ground truth object geometry to all methods for a fair comparison and focus the evaluation on hand reconstruction quality. We reserve optimizing for the object (shape recovery) to \cref{sec:ablation}. We retrain and evaluate all methods on HOI4D, DexYCB, and OakInk under this unified protocol. To validate the effectiveness of our prior, we plug \methodname{} into two representative baselines, EasyHOI and AlignSDF, as an object-conditioned hand prior that refines their hand results.

\begin{figure}[h]
    \centering
    \includegraphics[width=\linewidth]{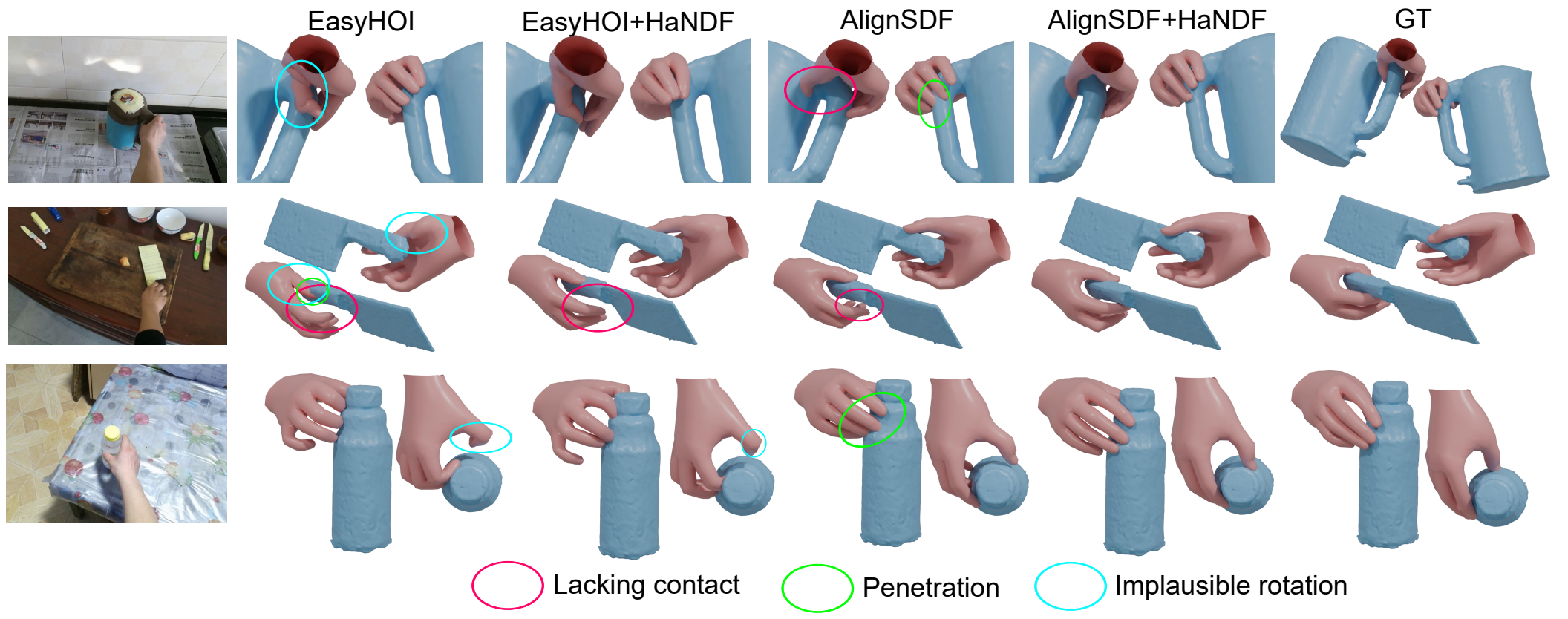}
    \caption{Qualitative results for in-the-wild hand recovery on HOI4D. Our HaNDF consistently improves the reconstruction quality of both EasyHOI and AlignSDF, alleviating common failure modes such as lacking contact, hand-object penetration, and implausible hand rotation.\vspace{-3mm}}
    \label{fig:qual-reconstruction}
\end{figure}

As shown quantitatively in \cref{tab:sota_comparison} and qualitatively in \cref{fig:qual-reconstruction}, integrating \methodname{} consistently improves two baselines across all three datasets. In particular, \methodname{} (AlignSDF-based) achieves the best MPVPE, MPJPE, and intersection volume on most benchmarks, demonstrating that our prior provides a strong and generalizable regularization for hand reconstruction under occlusion. We provide more in-the-wild hand recovery results in~\cref{{app:in_the_wild_hand_recov}}.

\vspace{-1mm}

\cref{tab:object_optim} compares hand-object recovery on the HOI4D test
set, using AlignSDF~\cite{chen2022alignsdf} as the baseline. Starting from
AlignSDF's predictions, applying HaNDF as a denoiser substantially improves
hand reconstruction quality, reducing $\Delta q$ from 14.77 to 9.60, MPJPE
from 12.08 to 11.74, and MPVPE from 11.90 to 11.52. This confirms that the
learned hand prior, conditioned on the predicted object, effectively
refines noisy hand outputs. We further refine the object mesh by retrieving
the nearest neighbor in the DeepSDF latent space (with PCA), which not
only improves object reconstruction (obj-F@2\% from 0.1187 to 0.1568) but also further boosts hand quality (MPJPE: 11.74 $\rightarrow$
11.50, MPVPE: 11.52 $\rightarrow$ 11.30). With additional object recovery (as shown in \cref{alg:object_retrieval}) with $K=8$ optimization, we observe better hand performance, with slight worse object metrics.
This experiment validates that
better object reconstruction directly translates into better hand
reconstruction, closing the loop between the two modalities.

\vspace{-3mm}
\begin{table}[h!]
  \centering
  \caption{\textbf{Hand-object recovery.} We evaluate different hand priors built upon the hand--object reconstruction results from AlignSDF baseline \cite{chen2022alignsdf} on the HOI4D dataset. We condition \methodname{} on the object predicted by AlignSDF. We also refine the object by retrieving the nearest neighbor in the object latent space (with PCA), possibly with further object recovery by optimization with different $K$, which show advanced performance. CD refers to Chamfer Distance. The best and second-best results are highlighted in \textbf{bold} and \underline{underlined text}, respectively.}
  \label{tab:object_optim}
  \resizebox{\linewidth}{!}{%
  \begin{tabular}{l|c|c|ccc|ccc}
  \toprule
  Method & Obj retrieval & Obj recovery & $\Delta q\downarrow$ & MPJPE$\downarrow$ & MPVPE$\downarrow$ & obj-CD$\downarrow$ & obj-F@1\%$\uparrow$ & obj-F@2\%$\uparrow$ \\
  \midrule
  AlignSDF~\cite{chen2022alignsdf} & \xmark & N/A & 14.77 & 12.08 & 11.90 & \underline{0.0909} & 0.0296 & 0.1187 \\
  \midrule
  \multirow{3}{*}{AlignSDF + \methodname{}} & \xmark & N/A & 9.60 & 11.74 & 11.52 & \underline{0.0909} & 0.0296 & 0.1187 \\
  & \checkmark & $K=1$ & \underline{9.59} & \underline{11.50} & \underline{11.30} & \textbf{0.0896} & \textbf{0.0449} & \textbf{0.1568} \\
  & \checkmark & $K=8$ & \textbf{9.57} & \textbf{11.45} & \textbf{11.25} & 0.1083 & \underline{0.0387} & \underline{0.1360} \\
  \bottomrule
\end{tabular}
  }
  \vspace{-4mm}
\end{table}

\begin{table}[th]
\centering
\setlength{\tabcolsep}{10pt}
\caption{Ablation over backbone, conditioning, and field geometry on HOI4D.}
\begin{tabular}{llccc}
\toprule
Backbone / conditioning & Field geometry & $\Delta_q\,(^\circ)$ & dNN$\,(^\circ)$ & m2m (mm) \\
\midrule
MLP, unconditional                & Euclidean             & 8.29  & 5.88 & 9.54 \\
MLP, unconditional                & Riemannian                & 7.16  & 6.52 & 8.60 \\
MLP + object condition            & Euclidean  & 8.15  & 5.12 & 9.06 \\
MLP + object condition            & Riemannian                 & 5.98  & \textbf{4.57} & 7.14 \\
Transformer, unconditional        & Riemannian                        & 6.10  & 5.75 & 7.40 \\
Transformer + object condition    & Euclidean  & 10.27 & 8.09 & 9.92 \\
Transformer + object condition    & Riemannian       & \textbf{5.52}  & 4.96 & \textbf{6.58} \\
\bottomrule
\end{tabular}
\vspace{-3mm}
\label{tab:ablation_ba_co_field}
\end{table}
\subsection{Ablation Studies}
\label{sec:ablation}

\noindent\textbf{Architecture}
We provide ablations on model architecture, conditioning, and the geometry of the neural field. As shown in~\cref{tab:ablation_ba_co_field}, Riemannian geometry achieves a large performance gain, and object conditioning also improves performance for both MLPs and Transformers. With object conditioning and a Riemannian field, the MLP-based model also achieves comparable performance.

\begin{wraptable}[9]{r}{0.6\linewidth}
  \centering
  \vspace{-4mm}
  \caption{%
    Object-conditioning ablation on hand denoising with noise level $\sigma{=}0.8$\,rad on HOI4D dataset.
  }
  \label{tab:obj_cond_ablation}
  \small
  \vspace{-3mm}
  \begin{tabular}{lccc}
    \toprule
    & $\Delta q$ ($^\circ$) $\downarrow$ & dNN ($^\circ$) $\downarrow$ & m2m $\downarrow$ \\
    \midrule
    Noisy input           & $11.74 $ & $12.43 $ & $11.13$ \\
    \midrule
    \methodname~w/ wrong condition & $5.80 $ & $5.25 $ & $7.05$ \\
    \methodname~w/o object token    & $6.10 $ & $5.75 $ & $7.40$ \\
    \methodname  & $\mathbf{5.52 }$ & $\mathbf{4.96 }$ & $\mathbf{	6.58}$ \\
    \bottomrule
  \end{tabular}
\end{wraptable}
\vspace{1mm}
\noindent\textbf{Object Conditioning}
To verify that \methodname{} indeed exploits the object as a meaningful condition rather than treating it as a nuisance input, we compare three variants in \cref{tab:obj_cond_ablation}. ``Wrong condition'' replaces the input object with a randomly sampled one from a different category, while ``w/o object token'' removes the object token from the model and retrains it with all other settings unchanged. 
Feeding a wrong object increases the joint-angle error, and removing the object token leads to an even larger drop in performance. These gaps confirm that \methodname{} effectively leverages the object condition to guide hand denoising.

\subsection{Runtime Analysis}
\vspace{-2mm}

\begin{wraptable}[11]{r}{0.6\linewidth}
\vspace{-4mm}
\centering
\small
\caption{Comparison of inference costs between the proposed HaNDF and LatentHOI~\cite{li2025latenthoi} for object-conditioned generation on HOI4D. Runtime is reported per sample. $^\dagger$With 200 steps, the applied step size is doubled.}
\label{tab:runtime_comparison}
\resizebox{\linewidth}{!}{
\begin{tabular}{lcccc}
\toprule
Method & \# Steps & FID $\downarrow$ & dNN $\downarrow$ & Runtime (ms) $\downarrow$ \\
\midrule
LatentHOI \cite{li2025latenthoi} & 1000 & 2.7843 & 6.41$^\circ$ & 64.5 \\
HaNDF & 200$^\dagger$ & 0.5482 & 5.07$^\circ$ & \textbf{12} \\
HaNDF & 500 & 0.5466 & 4.77$^\circ$ & 29 \\
HaNDF & 1000 & \textbf{0.5297} & \textbf{4.66$^\circ$} & 59 \\
\bottomrule
\end{tabular}
}
\end{wraptable}
To verify the runtime efficiency of the proposed HaNDF, we compare the time required to generate a single sample against LatentHOI \cite{li2025latenthoi} in~\cref{tab:runtime_comparison}. The experiment is done a single NVIDIA L40S GPU. Our proposed HaNDF (i) given the same iterations, can generate poses with substantially better FID and dNN scores faster than LatentHOI, and (ii) can match the LatentHOI metrics quicker, enabling less iterations and hence even quicker runtime.  We provide more details and analysis on the inference in~\cref{sec:inference_efficiency}

%% file: sec/5_conclusion.tex
\section{Conclusion}
\label{sec:conclusion}
\vspace{-1mm}

We presented HaNDF, an object-conditioned neural distance field for hand articulation on Riemannian manifolds. Coupled with manifold-aware optimization, the learned geometric prior supports denoising, completion, generation, and reconstruction. It can consistently improve robustness and physical plausibility across challenging benchmarks. More broadly, our results suggest that HaNDF provide a promising direction for learning structured interaction priors beyond Euclidean latent representations.

\paragraph{Limitations and future work}
\label{main_paper_future_work}
Our method relies on reasonably accurate object geometry and is not a true grasp prior as we do not model a hand-object joint NDF. Such an extension must account for the different geometries
of hand rotations and object latents, along with their
different noise scales and optimization step sizes, and may also need to incorporate hand–object contact dynamics. We will also extend our hand prior to bimanual interactions, as well as investigating more efficient training strategies. Finally, our model can be extended to the dynamic setting in a similar vein to NRMF~\cite{yu2025nrmf}. More future work analysis are shown in~\cref{app:future_work_append}.

\newpage
\begin{ack}
\vspace{-3mm}
T. Birdal was supported by a UKRI Future Leaders Fellowship
[grant number MR/Y018818/1]. The authors acknowledge support
from the UK AI Research Resource (AIRR Isambard AI) through
grant 0251-4584-0945-1 -- TopoFound.
\end{ack}

%% file: sec/6_appendix.tex
\clearpage
\appendix
\section*{Appendix}

\section{Riemannian Geometry}
\label{app:Riemannian}
Let $\Man$ be a smooth $m$-dimensional manifold embedded in an ambient Euclidean space $\Amb=\R^d$. For $\x\in\Man$, the \emph{tangent space} $T_{\x}\Man$ is the set of all velocity vectors of smooth curves passing through $\x$:
\begin{equation}
T_{\x}\Man
=
\left\{
\dot{\gamma}(0)
\;\middle|\;
\gamma:\R\to\Man,\;
\gamma(0)=\x,\;
\gamma \text{ smooth}
\right\}.\nonumber
\end{equation}
A \emph{Riemannian metric} $\mathbf{G}$ assigns every $\x\in\Man$ an inner product $\langle \ub,\vb\rangle_{\x} = \ub^\top \mathbf{G}_{\x} \vb$ for $\ub,\vb\in T_{\x}\Man$, varying smoothly with $\x$. The pair $(\Man,\mathbf{G})$ is a \emph{Riemannian manifold}. For a smooth $f:\Man\to\R$, the \emph{Riemannian gradient} $\grad\,f(\x)\in T_{\x}\Man$ is defined as the unique tangent vector satisfying
\begin{equation}
\mathrm{D}f(\x)[\vb]
=
\langle \vb,\grad\,f(\x)\rangle_{\x},
\qquad
\forall \vb\in T_{\x}\Man.
\end{equation}
For $\Man\subseteq \Amb$ and $f$ implemented for example by a neural network, automatic differentiation gives the Euclidean gradient $\nabla f(\x)\in\R^d$. The intrinsic gradient is obtained by projecting onto $T_{\x}\Man$:
\begin{equation}
\grad\,f(\x)=\Pi_{\x}(\nabla f(\x)),
\end{equation}
where $\Pi_{\x}:\R^d\to T_{\x}\Man$ is the orthogonal projection. This \emph{egrad-to-rgrad}~\cite{boumal2023introduction} conversion is the key enabler of our distance field operators.
A \emph{Riemannian gradient step} from $\x$ along a tangent vector $\boldsymbol{\xi}\in T_{\x}\Man$ is performed by mapping $\boldsymbol{\xi}$ back to the manifold, ideally via the \emph{exponential map} $\Exp_{\x}:T_{\x}\Man\to\Man$, yielding \emph{Riemannian gradient descent}~\cite{boumal2023introduction,birdal2019probabilistic} (RGD) of the form
\begin{equation}
\x_{t+1}
=
\Exp_{\x_t}
\left(
-\tau_t\,\grad\,f(\x_t)
\right),
\end{equation}
where $\tau_t>0$ is a step size, which can often be adaptively computed, \eg, via line search. 

\section{The Rotation Group $\SO$}
\label{sec:app:rotations}
For the sake of being self contained, we now present the space of 3D rotations as a Riemannian manifold, the \emph{special orthogonal group}:
\begin{equation}
\SO
=
\left\{
\Rot\in\R^{3\times 3}
\;\middle|\;
\Rot^\top \Rot=\mathbf{I},\;\det(\Rot)=1
\right\}.
\end{equation}
Its \emph{Lie algebra} is the vector space of skew-symmetric matrices:
\begin{equation}
\mathfrak{so}(3)
=
\left\{
\Skew\in\R^{3\times 3}
\;\middle|\;
\Skew^\top=-\Skew
\right\}.
\end{equation}

For $\Skew=(\omega_1,\omega_2,\omega_3)^\top\in\R^3$, define the hat operator $[\cdot]_\times$ so that $[\Skew]_\times \x=\Skew\times \x$ for all $\x\in\R^3$:
\begin{equation}
[\Skew]_\times
=
\begin{bmatrix}
0 & -\omega_3 & \omega_2 \\
\omega_3 & 0 & -\omega_1 \\
-\omega_2 & \omega_1 & 0
\end{bmatrix},
\end{equation}
The inverse map is denoted by $\mathrm{vee}(\cdot)$: $\mathrm{vee}([\Skew]_\times)=\Skew$.
The tangent space of $\SO$ at $\Rot$ is
\begin{equation}
T_{\Rot}\SO
=
\left\{
\Rot\Skew
\;\middle|\;
\Skew\in\mathfrak{so}(3)
\right\}.
\end{equation}
Indeed, for a smooth curve $\Rot(t)\in\SO$ with $\Rot(0)=\Rot$, differentiating the constraint $\Rot(t)^\top \Rot(t)=\mathbf{I}$ gives $\dot{\Rot}(0)^\top \Rot + \Rot^\top \dot{\Rot}(0)=\mathbf{0}$.
Thus $\Rot^\top\dot{\Rot}(0)$ is skew-symmetric, and every tangent vector can be written as $\dot{\Rot}=\Rot\Skew$ for some $\Skew\in\mathfrak{so}(3)$.
Using the Frobenius inner product $\langle \mathbf{A},\mathbf{B}\rangle_F = \mathrm{Tr}(\mathbf{A}^\top \mathbf{B})$, 
the Riemannian gradient of $f:\SO\to\R$ can be obtained from the Euclidean gradient $\nabla f(\Rot)$ by
\begin{equation}
\grad\,f(\Rot)
=
\Rot\,\skewrm\left(\Rot^\top\nabla f(\Rot)\right), \quad\text{where}\quad \skewrm\,(\mathbf{A})=\frac{1}{2}(\mathbf{A}-\mathbf{A}^\top).
\end{equation}
This projection removes the normal component and keeps only the component tangent to $\SO$.
The \emph{exponential map on $\SO$} is
\begin{equation}
\Exp_{\Rot}(\Rot\Skew)
=
\Rot\exp(\Skew),
\end{equation}
where $\exp(\Skew)$ is the matrix exponential. If $\Skew=[\Skew]_\times$ and $\vartheta=\|\Skew\|_2$, Rodrigues' formula gives
\begin{equation}
\exp(\Skew)
=
\mathbf{I}
+
\frac{\sin\vartheta}{\vartheta}\Skew
+
\frac{1-\cos\vartheta}{\vartheta^2}\Skew^2.
\end{equation}
The logarithmic map is the local inverse $\Log_{\Rot_1}(\Rot_2)=\Rot_1\log(\Rot_1^\top \Rot_2)$, where $\log(\cdot)$ is the matrix logarithm. The geodesic distance between rotations is
\begin{equation}
d_{\SO}(\Rot_1,\Rot_2)
=
\left\|
\log(\Rot_1^\top \Rot_2)
\right\|_{\mathfrak{so}(3)}
=
\cos^{-1}
\left(
\frac{\mathrm{Tr}(\Rot_1^\top \Rot_2)-1}{2}
\right).
\end{equation}
These matrix expressions are useful conceptually, but for articulated hands we use \emph{unit quaternions} because they provide a computationally convenient representation for gradient-based optimization.

\section{Implementation Details}
\label{app:implementation}

\subsection{Model Architecture}
\label{app:arch}

Our \methodname{} applies a transformer encoder that processes hand and object tokens jointly. The architecture details are:

\begin{enumerate}[nosep,leftmargin=*]
    \item Hand token projection: Linear layer $\mathbb{R}^4 \to \mathbb{R}^{256}$ for each quaternion
    \item Object token projection: Linear layer $\mathbb{R}^{64} \to \mathbb{R}^{256}$ for the single object token
    \item $[\mathrm{DIST}]$ token: Learnable embedding $\in \mathbb{R}^{256}$ to aggregrate information
    \item Positional encoding: Learnable positional embeddings for all tokens
\end{enumerate}

As illustrated in Fig. 9, HaNDF adopts a multi-head transformer as its backbone by default. The 15 hand tokens, the object token, and a learnable distance token are concatenated and jointly processed by the backbone. The output distance token is then fed into a lightweight head to predict the final distance.

\begin{figure}[h]
    \centering
    \includegraphics[width=0.7\linewidth]{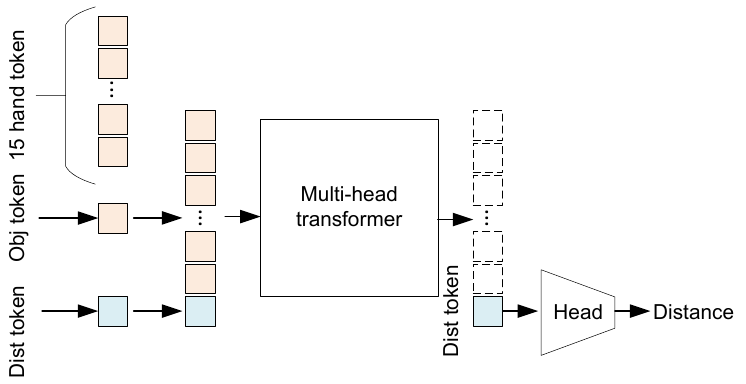}
    \caption{Diagram of HaNDF's transformer backbone, which combines hand, object, and distance tokens to predict the distance.}
    \label{fig:placeholder}
\end{figure}

MLPs are also applicable as the backbone network for HaNDF, offering faster convergence at the cost of slightly degraded performance (as shown in~\cref{tab:ablation_ba_co_field}). However, this performance gap does not necessarily preclude their applicability to downstream tasks, and in some cases MLP-based HaNDF can still achieve competitive results.

\subsection{Training Algorithm}

The algorithm to train HaNDF is as follow:

\begin{algorithm}[h]
\caption{Training HaNDF}
\label{alg:handf_training}
\begin{algorithmic}[1]

\Require Pose perturbation distribution $\mathcal{P}$,
dataset of objects and associated hand poses
$\mathcal{D}=\{(\mathcal{O}_j,\mathcal{D}_j)\}_{j=1}^{M}$,
where $\mathcal{D}_j=\{\pose_{j,n}\}_{n=1}^{N_j}$,
initial parameters $\Phi$ of $f_\Phi$

\Ensure Trained weights $\Phi$

\State Initialize an empty training set $\mathcal{D}^\prime \gets \emptyset$

\For{each object $\mathcal{O}_j$ in $\mathcal{D}$}
    \State Obtain object latent $\z_j$ by optimizing the DeepSDF autodecoder for 500 steps
    \State Add $(\pose_{j,n},\z_j,j)$ for all $n=1,\ldots,N_j$ to $\mathcal{D}^\prime$
\EndFor

\While{not converged}
    \For{each epoch}
        \State Sample random subset $\mathcal{D}^O$ from $\mathcal{D}^\prime$: $\mathcal{D}^O\subset\mathcal{D}^\prime$
        \State Perturb all hand poses in $\mathcal{D}^O$ 
        using the noise sampled from $\mathcal{P}$
         \State Update $\Phi$ by minimizing (using Adam)
         \[
         \sum\limits_{{(\pose^\prime,\z,j)}\in\mathcal{D}^O}
        \left\|
        f_\Phi(\pose^\prime\given\z)
        -
        \min_{\pose\in\mathcal{D}_j}
        d_{\Man_H}(\pose^\prime,\pose)
        \right\|
        \]
        
    \EndFor
\EndWhile

\end{algorithmic}
\end{algorithm}

The training dataset normally stores data in the form of sequences, where we first get the object-hand pairs for interacting actions in the dataset to build the training set. Specifically, for each object $\mathcal{O}_j$, we first fit DeepSDF to obtain the object code $z_j$ with 500 optimization steps, and we store the object code together with each hand pose under that object as a pair $(\pose_{j,n},\z_j,j)$, which serves as one training sample later. Note that for HOI4D and OakInk, we split the dataset based on objects, which means that the objects used during evaluation never appear during training. After constructing the training set, we sweep through the training samples to add perturbation noise based on an exponential distribution $\mathcal{P}$, and calculate the dNN between the noisy pose and all clean poses for that specific object to get the training label.

\subsection{Object Recovery}
\label{app:object_retrieval}

\cref{alg:object_retrieval} summarizes the object
recovery and hand refinement procedure. We first fit PCA to
the mean-centred object codes in training set and retain the
top-$d$ principal directions to define the retrieval space.
For each of the $K$ candidates, we initialize an object latent
code from $\mathcal N(\mathbf 0, 0.01^2\mathbf I)$ and optimize
it using the fixed DeepSDF decoder. Specifically, we minimize
the squared predicted SDF values at the input object surface
points, whose target SDF values are zero. We then project the
fitted code and the training codes into PCA space and retrieve
the nearest training code as our retrieved object code.

\begin{algorithm}[h]
\caption{Object recovery and hand refinement}
\label{alg:object_retrieval}
\begin{algorithmic}[1]
\Require Initial hand pose $\boldsymbol{\theta}^0$,
object surface points $\mathcal O^p$,
training latent codebook $\mathcal Z = \{\mathbf z_j\}_{j=1}^{M}$,
number of candidates $K$; PCA dimension $d$,
hand refinement steps $T$, candidate evaluation steps $T_c$, coded-shape DeepSDF $g_\psi$

\State Compute the top-$d$ principal directions $\mathbf V_d$
of the mean-centred codes in $\mathcal Z$

\For{$i=1,\ldots,K$}
    \State Initialize $\mathbf z_i^{(0)}
    \sim \mathcal N(\mathbf 0, 0.01^2\mathbf I)$

    \State Fit the object latent code from $\mathbf z_i^{(0)}$:
    \Statex \hspace{\algorithmicindent}
    $\displaystyle
    \hat{\mathbf z}_i \approx
    \arg\min_{\mathbf z}
    \frac{1}{|\mathcal O^p|}
    \sum_{\mathbf o \in \mathcal O^p}
    g_\psi(\mathbf z,\mathbf o)^2$

    \State Retrieve the nearest training code in PCA space:
    \Statex \hspace{\algorithmicindent}
    $\displaystyle
   \tilde{\mathbf z}_i  = \arg\min_{\z\in\mathcal{Z}}
    \left\|
    \mathbf V_d^\top
    (\hat{\mathbf z}_i - \mathbf z)
    \right\|_2$

    \State $\boldsymbol{\theta}_i^{(0)} \gets \boldsymbol{\theta}_0$
    \State Compute $\boldsymbol{\theta}_i^{(T_c)}$ by iterating Eq.~\eqref{eq:HaNDFGrad} $T_c$ times conditioned on $\tilde{\mathbf z}_i$

    \State $d_i^{(0)} \gets
    f_\Phi(\boldsymbol{\theta}_i^{(0)} \mid \tilde{\mathbf z}_i)$

    \State $d_i^{(T_c)} \gets
    f_\Phi(\boldsymbol{\theta}_i^{(T_c)} \mid \tilde{\mathbf z}_i)$
\EndFor

\State $\mathbf{s} \gets \mathrm{normalize}(\log(\mathbf{d}^{(0)}) - \mathrm{normalize}(\log(\mathbf{d}^{(T_C)}))$

\State $i^* \gets \arg\min_{i \in \{1,\ldots,K\}} s_i$

\State Compute $\boldsymbol{\theta}_{i^*}^{(T)}$ by iterating Eq.~\eqref{eq:HaNDFGrad} $T$ times conditioned on $\tilde{\mathbf z}_{i^*}$
\State \Return $\boldsymbol{\theta}_{i^*}^{(T)},\tilde{\mathbf z}_{i^*}$
\end{algorithmic}
\end{algorithm}

Starting from the initial hand pose, we apply $T_c$ Riemannian
updates conditioned on each retrieved object code. Our
candidate selection is motivated by the intuition that a
compatible object should require relatively little correction
of the initial hand pose. To evaluate this, we record the
predicted HaNDF distances before and after refinement. We
take their logarithms and standardize the initial and final
values separately across the $K$ candidates. We then select
the candidate object code with the smallest difference between its
standardized initial and final distances. We finally refine hand with this selected object code. Quantitative results are shown in~\cref{tab:object_optim}.

\section{Additional Results}
\label{app:results}

\subsection{Quantitative hand pose generation results}
\label{app:hand_generation_app}

We evaluate hand pose generation on the HOI4D dataset using FID, APD, and dNN. The reults on~\cref{tab:generation} corresponds to visualization in~\cref{fig:generation}
.

\begin{table}[h]
        \centering
        \captionof{table}{Quantitative evaluation of pose generation on HOI4D dataset.}
        \label{tab:generation}
        \small %
        \begin{tabular}{l|ccc}
            \toprule
            Model & FID $\downarrow$ & APD $\uparrow$ & dNN $\downarrow$  \\
            \midrule
            HMP~\cite{duran2024hmp} & 4.9338   & \textbf{44.72$^\circ$}  & 30.66$^\circ$ \\
            H-PoseNDF  & 2.685  & 21.48$^\circ$  & 30.8$^\circ$  \\
            H-NRDF & 0.7769  & 20.86$^\circ$  & 10.70$^\circ$  \\
            G-HOP~\cite{ye2024g} & 0.7316  & 9.34$^\circ$  & 10.45$^\circ$  \\
            LatentHOI~\cite{li2025latenthoi} & 2.7843 & 6.01$^\circ$ & 6.41$^\circ$ \\
            Text2HOI~\cite{cha2024text2hoi} & 7.05 & 31.15$^\circ$ & 26.85$^\circ$  \\
            \midrule
            \textbf{HaNDF (Ours)} & \textbf{0.5466} & 12.73$^\circ$ & \textbf{4.77$^\circ$}  \\
            \bottomrule
        \end{tabular}
\end{table}

As shown in~\cref{tab:generation}, HaNDF achieves the best FID (0.5466) and dNN (4.77$^\circ$) scores, indicating that our generated poses most closely match the real data distribution while maintaining high fidelity. Although HMP attains a higher APD, this comes at the cost of substantially worse FID and dNN, suggesting that its diversity stems from generating implausible poses rather than capturing the true pose manifold. Overall, HaNDF strikes the best balance between fidelity and diversity among all competing methods.

\subsection{More results on hand denoising}
\label{sec:hand_denoising_app}

\paragraph{Quantitative evaluation}
\cref{tab:hand_denoising} provides the per-$\sigma$ breakdown underlying \cref{fig:denoising}. HaNDF achieves the best $\Delta_q$ and $m2m$ at all three noise levels, with widening margins as $\sigma$ grows ($\Delta_q$: $3.67^\circ/5.52^\circ/8.00^\circ$; $m2m$: $4.38/6.58/10.08$), while its $dNN$ remains nearly constant ($4.74^\circ\!\to\!5.20^\circ$) and becomes the lowest from $\sigma\!\geq\!0.8$. H-PoseNDF's marginally lower $d_{\mathrm{NN}}$ at $\sigma\!=\!0.4$ ($3.81^\circ$) is offset by substantially worse $\Delta_q$ ($5.67^\circ$) and $m2m$ ($6.62$), evidencing mode collapse rather than better denoising. HMP trails all distance-field methods across every metric.

\begin{table}[h]
\centering
\caption{Quantitative evaluation of hand denoising on HOI4D dataset across three representative noise levels. All methods receive identical noisy inputs.}
\label{tab:hand_denoising}
\setlength{\tabcolsep}{4pt}
\begin{tabular}{l|ccc|ccc|ccc}
\toprule
& \multicolumn{3}{c|}{$\sigma = 0.4$ rad} & \multicolumn{3}{c|}{$\sigma = 0.8$ rad} & \multicolumn{3}{c}{$\sigma = 1.5$ rad} \\
Model & dNN$\downarrow$ & $\Delta_q$ $\downarrow$ & m2m $\downarrow$
      & dNN$\downarrow$ & $\Delta_q$ $\downarrow$ & m2m $\downarrow$
      & dNN$\downarrow$ & $\Delta_q$ $\downarrow$ & m2m $\downarrow$ \\
\midrule
HMP~{\scriptsize\citep{duran2024hmp}}   & 7.29$^\circ$ &  5.33$^\circ$ &  5.90 & 11.23$^\circ$ & 10.26$^\circ$ & 10.37 & 17.64$^\circ$ & 17.76$^\circ$ & 16.38 \\
H-PoseNDF                 & \textbf{3.81}$^\circ$ &  5.67$^\circ$ &  6.62 &  5.88$^\circ$ &  8.29$^\circ$ &  9.54 & 10.97$^\circ$ & 14.10$^\circ$ & 15.59 \\
H-NRDF                    & 5.57$^\circ$ &  4.86$^\circ$ &  5.89 &  6.52$^\circ$ &  7.16$^\circ$ &  8.60 &  7.66$^\circ$ & 10.15$^\circ$ & 12.49 \\
\methodname               & 4.74$^\circ$ & \textbf{3.67}$^\circ$ & \textbf{4.38} & \textbf{4.96}$^\circ$ & \textbf{5.52}$^\circ$ & \textbf{6.58} & \textbf{5.20}$^\circ$ & \textbf{8.00}$^\circ$ & \textbf{10.08} \\
\bottomrule
\end{tabular}
\vspace{-2mm}
\end{table}

In addition to the quantitative metrics presented in \cref{tab:hand_denoising}, we provide extended qualitative evaluations of our pose denoising method. Due to space constraints in the main manuscript, we present the remainder of our visual results here. 

\begin{figure}[h]
    \centering
    \includegraphics[width=\linewidth]{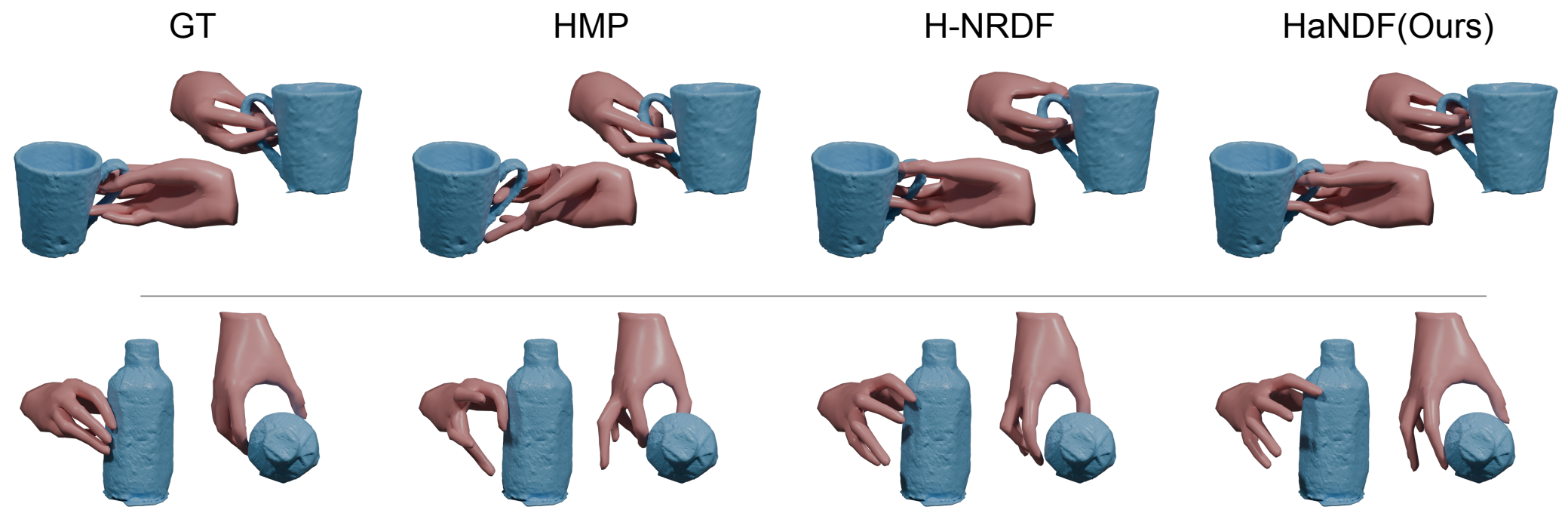}
    \caption{Extended qualitative comparison of pose denoising across different methods on HOI4D. We provide further comparison cases to supplement the results in the main paper. \methodname~(ours) is evaluated against HMP and H-NRDF, with Ground Truth (GT) provided for reference.}
    \label{fig:supp_qual_denoising}
\end{figure}

\paragraph{More qualitative results}
\cref{fig:supp_qual_denoising} showcases further examples of the iterative denoising process. Starting from highly corrupted initial states, it is evident that over $1000$ steps, HaNDF successfully resolves penetrations and guides the hand toward physically plausible, tight grasps across diverse object topologies. 

\begin{figure}[h]
    \centering
    \includegraphics[width=\linewidth]{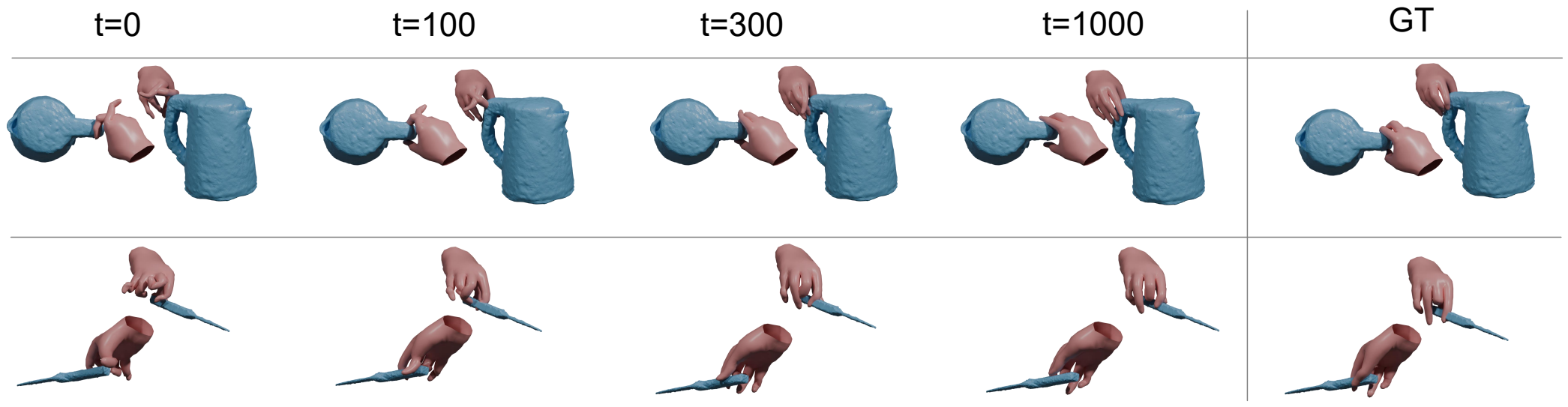}
    \caption{Additional qualitative pose denoising iterations for hand-object interaction on HOI4D. We show two further examples of the refinement steps from $t=0$ to $t=1000$ , with Ground Truth (GT) provided for reference.}
    \label{fig:supp_qual_denoising_process}
\end{figure}

Furthermore, \cref{fig:supp_qual_denoising_process} provides additional baseline comparisons for the denoising task. Consistent with our main findings, our method produces hand-object configurations that align more accurately with the ground truth manifold, avoiding the artifacts observed in HMP and H-NRDF.

\paragraph{Robustness under controlled object-geometry perturbations}

We have also studied about robustness under controlled object-geometry perturbations. For each test object we sample 8,192 surface points on the mesh, apply a controlled corruption, re-fit the DeepSDF latent, and re-run denoising experiments. Entries are the cost of the corruption measured against the same fit on uncorrupted points. We evaluate robustness to three types of perturbations: point-wise noise at different noise levels $\sigma_p$, partial visibility, and point-cloud sparsity, and the results are shown in \cref{tab:object_robustness}.

\begin{table}[h]
\centering
\small
\caption{
Robustness to object-input perturbations. We evaluate robustness to three types of perturbations: point-wise noise at different noise levels $\sigma_p$, partial visibility, and point-cloud sparsity. The results demonstrate good robustness across these perturbations.}
\label{tab:object_robustness}
\begin{tabular}{llccc}
\toprule
Object input & Setting
& $+\Delta_q$ ($^\circ$)
& $+\mathrm{dNN}$ ($^\circ$)
& $+\mathrm{m2m}$ (mm) \\
\midrule
Point noise $\sigma_p$
& 0.01 & +0.00 & +0.00 & +0.02 \\
& 0.03 & +0.02 & +0.06 & +0.07 \\
& 0.05 & +0.05 & +0.06 & +0.09 \\
\addlinespace
Partial visibility
& 25\% occluded & +0.03 & +0.05 & +0.06 \\
& 50\% occluded & +0.04 & +0.07 & +0.10 \\
\addlinespace
Point-cloud sparsity
& 75\% subsample (control) & +0.00 & +0.00 & +0.01 \\
\bottomrule
\end{tabular}
\end{table}

Conditioning is robust to corrupted object inputs. Point noise at $\sigma_p = 0.05$ costs $0.05^\circ$ of $\Delta_{q_r}$ and occluding half the object costs $0.04^\circ$. Fits from occluded inputs land at cosine distance $0.63$ from the clean fits, which is more than half the way to a different object. This shows the prior is insensitive to where the latent moves within a neighbourhood.

\subsection{More results on in-the-wild hand recovery}
\label{app:in_the_wild_hand_recov}

\cref{tab:sota_comparison} shows results under GT object code to focus on the hand performance. Here we further provide the EasyHOI-based end-to-end experiment by using EasyHOI-predicted objects to match to the nearest training object by shape, and then conditioning its latent on HaNDF to do the hand refinement.

\begin{table}[h]
  \centering
  \caption{EasyHOI-based end-to-end experiment. We use EasyHOI-predicted objects to match to the nearest training object by shape, and then conditioning its latent on HaNDF to do the hand refinement.}
  \label{tab:easyhoi_in_the_wild}
  \small
  \begin{tabular}{lcccc}
    \toprule
    Method & $\Delta_q$($^\circ$) & MPVPE (mm) & MPJPE (mm) & IntVol ($\mathrm{cm}^3$) \\
    \midrule
    EasyHOI & 20.82 & 16.64 & 17.06 & 13.28 \\
    EasyHOI + HaNDF & 18.60 & 15.49 & 15.88 & 8.11 \\
    \midrule
    Improvement (\%) & $+10.6$ & $+6.9$ & $+6.9$ & $+38.9$ \\
    \bottomrule
  \end{tabular}
\end{table}

As shown in \cref{tab:easyhoi_in_the_wild}, the HaNDF performs reasonably well under end-to-end setting of in-the-wild recovery.

\subsection{Comparison of object representations}
\label{app:resultsObjects}
In order to validate the efficiency of the applied DeepSDF-style auto-decoder for obtaining the latent code, we compare it with other object encoding methods, including COD-VAE~\cite{cho2025representing} and Tencent-XR-3DGen~\cite{yang2025pandora3d} (based on 3DShape2VecSet~\cite{zhang20233dshape2vecset}). The DeepSDF-style
auto-decoder generates
a single 64-dimensional vector, compared to $32{\times}32$ tokens for
COD-VAE and $2048{\times}64$ for Tencent-XR-3DGen. Note that we retrain the HaNDF model under the same settings for each object encoder to keep the comparison fair.

\begin{table}[H]
  \centering
  \small
  \caption{%
    Ablation on object conditioning representations. We compare three pretrained object encoders as the source of object tokens fed into \methodname{} on HOI4D.
  }
  \label{tab:obj_encoder_ablation}
  \begin{tabular}{lccc}
    \toprule
     Object encoder & $\Delta q$ $\downarrow$ & dNN $\downarrow$ & m2m (mm) $\downarrow$ \\
    \midrule
    \methodname + COD-VAE~\cite{cho2025representing}                      & 6.11$^\circ$  & 5.68$^\circ$ & 7.46 \\
    \methodname + Tencent-XR-3DGen~\cite{yang2025pandora3d}             & 5.70$^\circ$ & 5.17$^\circ$ & 6.78 \\
    \methodname + DeepSDF-style auto-decoder             &$\mathbf{5.52}^\circ$  &$\mathbf{4.96}^\circ$ & $\mathbf{6.58}$  \\
    \bottomrule
  \end{tabular}
\end{table}

As shown in \cref{tab:obj_encoder_ablation}, the DeepSDF-style
auto-decoder consistently outperforms both COD-VAE and
Tencent-XR-3DGen, despite producing by far the most compact latent. Two factors account
for this gap. First, the latter two are point-cloud-based encoders
whose outputs are permutation-equivariant. Therefore, the point-cloud encodings carry
no canonical ordering across instances, so the downstream transformer
is hard to rely on positional cues to align semantically corresponding
regions. Although we use the mean as the aggregation function for Tencent-XR-3DGen, it still underperforms compared to the applied DeepSDF-style auto-decoder. Second, the DeepSDF latent is optimized
end-to-end to reconstruct the object's signed distance field,
producing a compact, semantically aligned code in which geometrically
similar shapes occupy nearby regions of the latent space. When the conditioning
signal is too unstructured to be consumed effectively, \methodname{}
degenerates toward an unconditional hand prior, mirroring the
``w/o object token'' and ``wrong condition'' rows of
\cref{tab:obj_cond_ablation}.

\subsection{Transformer architecture Ablation}
We ablate three key design choices of our transformer architecture in
\cref{tab:ablation_arch} with depth and the granularity of the object
condition.
Our final configuration uses 4 layers, a single
object token, and a dedicated distance token, which shows the better performance among different settings.

\begin{table}[h]
  \centering
  \caption{Ablation on transformer architecture design choices on HOI4D dataset.}
  \label{tab:ablation_arch}
  \begin{tabular}{lccccc}
    \toprule
    Variant & $\Delta_q$  $\downarrow$ & dNN $\downarrow$ & Tokens(N) &  Params & FLOPs (G) \\
    \midrule
    2 layers 
    & 4.95$^\circ$ & 6.08$^\circ$  &17	 &1.63& 0.027  \\
    4 layers & 5.52$^\circ$ & 4.96$^\circ$ &17	&3.21  & 0.054 \\
    8 layers & 5.73$^\circ$ & 5.07$^\circ$  &17&6.36	& 0.109 \\
    \midrule
    1 object tokens & \textbf{5.52}$^\circ$ & \textbf{4.96}$^\circ$ &17	&3.21  & 0.054  \\
    64 object tokens & 6.40$^\circ$  &  6.09$^\circ$&80 & 3.21	& 0.266\\
    \bottomrule
  \end{tabular}
\end{table}

\paragraph{Depth}
Increasing depth from 2 to 4 layers trades a small regression in
$\Delta_q$ ($4.95^\circ \!\to\! 5.52^\circ$) for a substantial
improvement in dNN ($6.08^\circ \!\to\! 4.96^\circ$), indicating
that the shallower model fits per-joint quaternions slightly better
but generalizes worse in the geodesic neighborhood sense. Pushing
further to 8 layers degrades both metrics ($5.73^\circ$ / $5.07^\circ$)
while doubling parameters and FLOPs. Given the relatively small input
set (17 tokens) and the bounded complexity of the per-joint geodesic
field, 4 layers offer the best accuracy/compute trade-off; deeper
models appear to overfit without additional capacity benefit.

\paragraph{Object-token granularity}
We compare encoding the 64-d DeepSDF object latent as a single
condition token (a learned linear projection to the transformer width)
against spreading it over 64 tokens (one per latent dimension).
Despite the $5\times$ increase in attention FLOPs (0.054\,G\,$\rightarrow$\,0.266\,G),
the 64-token variant is markedly worse ($\Delta_q$: $5.52^\circ \rightarrow 6.40^\circ$).
Splitting the 64-dimensional object code into 64 separate tokens breaks up the shape information, so the transformer must combine these tokens through attention. Using a single token keeps the full object code together and allows each hand-joint token to access it directly.

\paragraph{Results across random seeds}
We repeat the comparison with and without object optimization
over five random seeds. As shown in~\cref{tab:multiseed_reconstruction}, object optimization
reduces MPJPE by $0.236 \pm 0.005$ mm and MPVPE by
$0.216 \pm 0.005$ mm, showing consistent improvements.

\begin{table}[h]
\centering
\caption{Hand reconstruction results for AlignSDF + HaNDF
over five random seeds on HOI4D. The gain is computed as the error without object
optimization minus the error with object optimization.}
\label{tab:multiseed_reconstruction}
\begin{tabular}{lccc}
\toprule
Metric & Without obj. optim. & With obj. optim. & Gain \\
\midrule
$\Delta_q\,(^\circ)$
& $9.5985 \pm 0.0008$
& $9.5920 \pm 0.0011$
& $0.0065 \pm 0.0011$ \\
MPJPE
& $11.735 \pm 0.002$
& $11.499 \pm 0.006$
& $0.236 \pm 0.005$ \\
MPVPE
& $11.516 \pm 0.002$
& $11.300 \pm 0.005$
& $0.216 \pm 0.005$ \\
\bottomrule
\end{tabular}
\end{table}

\subsection{Ablation of PCA Dimension on Object Retrieval}

During inference, the reconstructed object latent is projected into this PCA subspace, and nearest-neighbor retrieval is performed in the projected space. The retrieved latent is then used for the subsequent hand projection. 
Therefore, we also evaluate the effect of the PCA dimension used for
nearest neighbor retrieval. As shown in~\cref{tab:pca_dimension}, reconstruction performance
improves as the dimension increases from 8 to 32 and remains
similar beyond 32 dimensions. The default 32-dimensional
setting achieves nearly identical performance to retrieval
in the full latent space, indicating that dimensionality reduction has a negligible effect on retrieval quality.

\begin{table}[h]
\centering
\caption{Effect of the PCA dimension on object retrieval on HOI4D. The 32-dimensional setting is used in~\cref{tab:object_optim}. Note that 64 dimensions correspond to the full latent space.\vspace{2mm}}
\label{tab:pca_dimension}
\begin{tabular}{lccc}
\toprule
PCA dimension $K$ & $\Delta_q\,(^\circ)\downarrow$
& MPJPE (mm) $\downarrow$ & MPVPE (mm) $\downarrow$ \\
\midrule
2                & 9.68 & 11.73 & 11.53 \\
8                & 9.74 & 11.70 & 11.50 \\
16               & 9.63 & 11.63 & 11.42 \\
32 (default)     & \textbf{9.59} & \textbf{11.50} & \textbf{11.30} \\
48               & 9.60 & 11.53 & 11.33 \\
64 (full latent) & 9.59 & 11.52 & 11.32 \\
\bottomrule
\end{tabular}
\end{table}

One advantage for introducing PCA is scalability. As the object latent library grows, retrieval in the reduced space substantially lowers the search cost while preserving the optimization performance shown above. Since HaNDF is designed to retrieve the best candidate from an object library rather than generalize to unseen object categories, this provides an efficient implementation for large-scale retrieval.

\subsection{Inference Efficiency}
\label{sec:inference_efficiency}

We evaluate the effect of the number of projection
iterations on denoising accuracy and runtime.
We use 3,000 poses with noise level $\sigma=0.8$ and
step size $\alpha=0.02$, measuring runtime on a single
NVIDIA L40S GPU at a batch size of 128.
As shown in~\cref{tab:iteration_cost}, 200 iterations
already provide accuracy comparable to the default
500 iterations, while reducing the cost from 29 to
12 ms per pose. Increasing the iteration count beyond
500 provides no further improvement in these metrics.

\begin{table}[h]
\centering
\caption{Denoising accuracy and runtime for different numbers
of Riemannian projection iterations. The default setting
uses 500 iterations.\vspace{2mm}}
\label{tab:iteration_cost}
\begin{tabular}{lcccccccc}
\toprule
Iterations
& 10 & 50 & 100 & 200 & 350 & 500 & 750 & 1000 \\
\midrule
$\Delta_q\,(^\circ)\downarrow$
& 10.18 & 6.97 & 5.86 & \textbf{5.48} & 5.49 & 5.52 & 5.56 & 5.57 \\
$d_{\mathrm{NN}}\,(^\circ)\downarrow$
& 10.88 & 7.44 & 5.95 & 5.16 & 4.97 & \textbf{4.96} & 4.98 & 4.98 \\
Time (ms/pose) $\downarrow$
& \textbf{0.6} & 3 & 6 & 12 & 21 & 29 & 44 & 59 \\
\bottomrule
\end{tabular}
\end{table}

\subsection{Generalization to Unseen Object Categories}
\label{app:generalization}
We conduct a category-shift experiment as an additional analysis to examine the behavior of the learned prior on object categories that are entirely absent from training. Specifically, we evaluate our HOI4D-trained model on three OakInk categories with no counterpart among the HOI4D categories, including sprayer, lotion pump, and camera. The test set includes 4 objects with 2,990 poses.

\begin{table}[H]
    \centering
    \caption{Comparison of different models and settings.\vspace{2mm}}
    \label{tab:model_comparison_category}
    \begin{tabular}{lccc}
        \toprule
        \textbf{Model / Setting}
        & $\Delta_q(\hat{\circ})\downarrow$
        & $d_{\mathrm{NN}}(\hat{\circ})\downarrow$
        & $\mathrm{m2m}(\mathrm{mm})\downarrow$ \\
        \midrule
        Noisy input
        & 11.78 & 11.46 & 11.25 \\
        HOI4D-HaNDF (unseen category)
        & 8.67 & 7.50 & 9.28 \\
        OakInk-HaNDF (in-category)
        & \textbf{5.01} & \textbf{3.77} & \textbf{6.58} \\
        \bottomrule
    \end{tabular}
\end{table}

As shown in~\cref{tab:model_comparison_category}, the HOI4D-trained model improves all three metrics without adaptation, reducing $\Delta_q$ from $11.78^\circ$ to $8.67^\circ$, $d_{\mathrm{NN}}$ from $11.46^\circ$ to $7.50^\circ$, and $\mathrm{m2m}$ from $11.25$ to $9.28$~mm. These results suggest that the learned prior can improve hand poses even for object categories absent from training. However, a gap remains relative to the OakInk-trained model, indicating that category-specific training remains beneficial.

\begin{figure}[h]
    \centering
    \includegraphics[width=\linewidth]{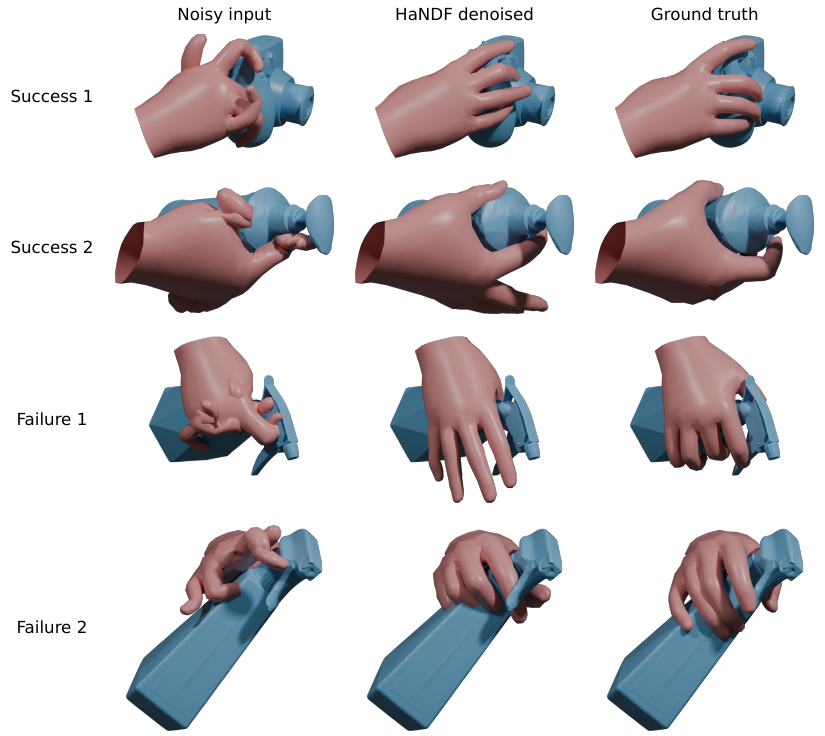}
    \caption{Generalization to Unseen Object Categories (by testing HOI4D trained HaNDF on OakInk unseen Categories).}
    \label{fig:HaNDF_OakInk_Success_Failure}
\end{figure}

\cref{fig:HaNDF_OakInk_Success_Failure} shows qualitative examples of unseen object categories. In the successful cases, HaNDF corrects distorted finger configurations and brings the hand pose closer to the ground truth. The failure cases illustrate remaining limitations: fingers may remain overextended or fail to recover the reference grasp, even when some of the initial pose errors are corrected.

\section{Baselines}
\label{app:baselines}
Additional information on our baselines can be found below. We compare \methodname{} against baselines spanning three categories:
\begin{enumerate}[nosep,leftmargin=*]
    \item Hand-domain adaptations of pose distance fields: PoseNDF~\cite{tiwari2022pose} and NRDF~\cite{he2024nrdf} are
distance-field priors originally proposed for full-body SMPL pose.
Since no public hand counterpart exists, we re-implement both on the
15-joint MANO manifold and retrain them on hand-object datasets, denoted
H-PoseNDF and H-NRDF. H-PoseNDF performs Euclidean
gradient descent with per-step quaternion re-normalization, while
H-NRDF follows the Riemannian formulation on $(\mathbb{S}^3)^{15}$.
Both are unconditional (no object input), serving as the most direct
ablations of \methodname{}'s object conditioning.
\item Hand / hand--object generative priors:
HMP~\cite{duran2024hmp} is a learned unconditional hand-pose
prior. G-HOP~\cite{ye2024g}, LatentHOI~\cite{li2025latenthoi},
and Text2HOI~\cite{cha2024text2hoi} are recent
hand-object generative models that condition on object geometry
(and text for Text2HOI). These cover the current state of the arts in
object-conditioned hand pose generation. For motion generation methods (Text2HOI~\cite{cha2024text2hoi} and LatentHOI~\cite{li2025latenthoi}), we adapt them to the single-frame right-hand pose setting
and retrain both models for a fair comparison. 
\item Single-image hand--object reconstruction:
For the reconstruction task we additionally compare with
EasyHOI~\cite{liu2025easyhoi},
AlignSDF~\cite{chen2022alignsdf}, and
IHoI~\cite{ye2022s}. To isolate the
contribution of the hand prior from heterogeneous object pipelines,
we feed ground-truth object geometry to all methods and evaluate hand
reconstruction quality. \methodname{} is plugged into EasyHOI and
AlignSDF as an object-conditioned refinement prior on top of their
pipelines.
\end{enumerate}

\section{Datasets}
\label{app:datasets}
Our experiments use three hand-object interaction datasets: HOI4D~\cite{liu2022hoi4d}, DexYCB~\cite{chao2021dexycb}, and OakInk~\cite{yang2022oakink}. 
We also use the MANO hand model~\cite{mano} to represent hand.
\begin{enumerate}[nosep,leftmargin=*]
    \item HOI4D~\cite{liu2022hoi4d} is a large-scale 4D egocentric dataset for category-level human-object interaction, containing approximately 2.4M RGB-D frames across 4{,}000+ sequences with 800 object instances, recorded at $1920{\times}1080$ with calibrated intrinsics. Each frame is densely annotated with MANO hand parameters, 6-DoF object poses, action labels, and 2D/3D segmentation. Following an object-based split, we focus on 6 rigid categories: instances N01--N40 (219 objects, 63K poses across 1{,}061 sequences) are used for training and the unseen instances N46--N50 (30 objects, 7.5K poses across 123 sequences) for testing, ensuring no object overlap between splits. We further apply contact-based filtering using the action annotations to retain only frames in which the hand physically interacts with the object.
\item 
DexYCB~\cite{chao2021dexycb} is a large-scale benchmark of third-person hand grasping, comprising 1{,}000 grasping sequences performed by 10 subjects on 21 YCB objects. We adopt the standard \texttt{s0} protocol, yielding approximately 582K annotated frames. Each frame provides MANO pose and shape parameters, 6-DoF object poses, and pixel-level segmentation.
\item
OakInk~\cite{yang2022oakink} is a large-scale hand-object interaction repository. We use its image-based subset, OakInk-Image, to validate our HaNDF. This subset contains ground-truth 3D MANO and object pose annotations. We use sequences from 7 object categories to build our training and test sets, with 53 objects (38K poses) for training and 14 unseen objects (10K poses) for testing.
\end{enumerate}

\section{Evaluation Metrics}
\label{app:metrics}
We now provide further details of the evaluation metrics. 

\begin{enumerate}[itemsep=0.1em,leftmargin=*,topsep=0.1em]
    \item dNN (distance to the nearest neighbor). For a predicted pose, dNN is the root-mean-square per-joint geodesic distance over the 15 finger joints between a pose and its nearest neighbor among the poses of the same object category in the training set. A small dNN indicates that the
grasp lies close to the valid-pose manifold. dNN specifically penalizes off-manifold
artifacts (such as twisted joints and implausible flexion).
\item $\Delta_q$ (quaternion geodesic distance to ground truth).
$\Delta_q$ is the root-mean-square per-joint geodesic error between the predicted and ground-truth quaternions over the evaluated joint set (all 15 finger joints for denoising, and only the masked joints for
completion).
\item FID (distribution fidelity). Following \cite{he2024nrdf},
we fit Gaussian distributions to the joint-angle features of the
predicted and reference poses. FID is the Fr\'echet distance between
these distributions, measuring differences in their means and
covariances.
\item APD (average pairwise distance).
For generation and completion, we draw $K=10$ samples per condition
and compute their average pairwise geodesic distance on
$(\mathbb{S}^3)^{15}$. Higher APD indicates greater diversity
among the generated grasps.
\item m2m (marker-to-marker distance).
Errors in proximal joint rotations can shift from several downstream
joints to the whole hand. To capture this effect, we also report m2m (marker-to-marker distance), the mean
Euclidean distance between the predicted and ground-truth hand vertices.
\item MPJPE / MPVPE (3D reconstruction error).
For single-image hand--object recovery we report the standard
Mean Per-Joint and Mean Per-Vertex Position Errors between the
predicted and ground-truth hand.
\item I.V. (interaction volume).
I.V. (cm$^3$) is the volume of the intersection between the
predicted hand mesh and the object mesh, computed by voxelizing both
meshes at $1\,$mm resolution. It quantifies physical implausibility:
a low MPJPE coupled with a high I.V. indicates a pose that is
numerically close to GT but penetrates the object.
\item Object reconstruction metrics.
For the object recovery experiment (\cref{tab:object_optim}),
we independently scale each predicted and ground-truth mesh so
that the longest half-side of its bounding box is one. We then
report the symmetric Chamfer distance (obj-CD) and F-score
(obj-F@$\tau$) at thresholds $\tau \in \{0.01, 0.02\}$ in
these normalized units.
\end{enumerate}

\section{Future Work}
\label{app:future_work_append}

As mentioned in \cref{main_paper_future_work}, our future work will focus on joint hand-object distance field, motion prior extension and bimanual interaction.

As a preliminary study of the bimanual interactions, we provide initial results on  OakInk2 \cite{zhan2024oakink2}, which annotates both left and right hand together with the subjects. We filter OakInk2 frames where both hands contact the same rigid object. It gives $30417$ train poses over 30 objects and $3441$ test poses over 5 held-out instances, split by object ID.

We refine the two-hand predictions of HaMeR~\cite{pavlakos2024reconstructing} on the held-out objects using HaNDF to demonstrate its applicability to the bimanual setting. The results are shown in~\cref{tab:bimanual}. To enable refinement of both hands, we increase the number of hand tokens in our transformer from $15$ to $30$ and retrain the prior.

\begin{table}[h]
\centering
\caption{Effect of hand priors on bimanual hand refinement using the OakInk2 dataset. Results are reported for the left hand, right hand, and both hands. The best results are shown in bold.\vspace{2mm}}
\label{tab:bimanual}
\begin{tabular}{lccc}
\toprule
Method & Left & Right & Both-hand mean \\
\midrule
HaMeR
& 10.00 / 9.84 & 10.38 / 10.38 & 10.19 / 10.11 \\
+ right-hand-only prior (15 joints)
& -- & \textbf{9.75 / 9.77} & -- \\
+ bimanual prior (30 joints)
& \textbf{9.13 / 8.98} & 9.97 / 9.99 & \textbf{9.55 / 9.49} \\
\bottomrule
\end{tabular}
\end{table}

We observe that the bimanual results on right hand underperform the right-hand only model. We provide the follow explaination: Firstly, both the bimanual prior and the right-hand-only prior are trained from the same training set. However, for the right-hand-only model, we additionally mirror the left hand and use it as augmentation, effectively providing additional training examples for the right-hand prior.
Secondly, adding additional hand increases the dimensionality of the optimization space, which may make projection and optimization more difficult.
Thirdly, since our HaNDF is trained only on the right hand, so its zero-level set was specialized only through object conditioning.

In terms of the joint-field for hand-object interaction, a naive extension is denoising hand pose and object latent jointly. We found it is affected by two main phenomenon:
The first is that the heterogeneity of manifolds: hand articulation sits on a compact product of spheres with a well-behaved geodesic structure, while DeepSDF latents live in an unbounded Euclidean space whose plausible region is defined only implicitly, by the distribution of the training latents. A single joint objective therefore has to commit to two different notions of distance at once.
The second is a matter of scale. The amount of hand-pose noise and the amount of latent noise that correspond to a comparable degree of semantic corruption differ by orders of magnitude, so a shared step size is either too coarse for one field or too timid for the other. We therefore regard a properly normalized joint field our next-step work.

Second, we envision that neural distance fields are applicable to cross-embodiment scenarios~\cite{zhang2026samor} where different skeletons are jointly represented. 
We also like to extend our work to factor in motion and dynamics similar to and beyond~\cite{he2026molingo,rempe2021humor,yu2025nrmf,yu2024signavatars}, and operate in multi-view settings~\cite{wang2026dispose}.

%% file: checklist.tex
\section*{NeurIPS Paper Checklist}

\begin{enumerate}

\item {\bf Claims}
    \item[] Question: Do the main claims made in the abstract and introduction accurately reflect the paper's contributions and scope?
    \item[] Answer: \answerYes{}
    \item[] Justification: The abstract and introduction state that HaNDF is an object-conditioned neural distance field for hand articulation, with applications to generation, denoising, completion, and reconstruction. These claims are supported by the method formulation and experiments

\item {\bf Limitations}
    \item[] Question: Does the paper discuss the limitations of the work performed by the authors?
    \item[] Answer: \answerYes{}
    \item[] Justification: The paper includes a limitations and future work paragraph in \cref{sec:conclusion}, noting that the method relies on reasonably accurate object geometry and does not yet model a full joint hand-object neural distance field.

\item {\bf Theory assumptions and proofs}
    \item[] Question: For each theoretical result, does the paper provide the full set of assumptions and a complete (and correct) proof?
    \item[] Answer: \answerNA{}
    \item[] Justification: The paper does not present new formal theorems requiring proofs. The mathematical sections define the Riemannian geometry, quaternion manifold, geodesic distances, and optimization operators used by the method.


    \item {\bf Experimental result reproducibility}
    \item[] Question: Does the paper fully disclose all the information needed to reproduce the main experimental results of the paper to the extent that it affects the main claims and/or conclusions of the paper (regardless of whether the code and data are provided or not)?
    \item[] Answer: \answerYes{}
    \item[] Justification: The paper describes the evaluated datasets, baselines, metrics, downstream tasks, optimization settings, architecture, and training configuration. Additional implementation details are provided in the appendix.


\item {\bf Open access to data and code}
    \item[] Question: Does the paper provide open access to the data and code, with sufficient instructions to faithfully reproduce the main experimental results, as described in supplemental material?
    \item[] Answer: \answerYes{}
    \item[] Justification: The implementation has been publicly released, along with instructions for environment setup and reproducing the main experimental results. All experiments use publicly available datasets.

\item {\bf Experimental setting/details}
    \item[] Question: Does the paper specify all the training and test details (e.g., data splits, hyperparameters, how they were chosen, type of optimizer) necessary to understand the results?
    \item[] Answer: \answerYes{}
    \item[] Justification: The paper specifies dataset splits, baselines, evaluation metrics, downstream task protocols, optimization iterations and step sizes, model architecture, optimizer, learning rate, batch size, number of epochs, noise sampling, and hardware.

\item {\bf Experiment statistical significance}
    \item[] Question: Does the paper report error bars suitably and correctly defined or other appropriate information about the statistical significance of the experiments?
    \item[] Answer: \answerNo{}
    \item[] Justification: The paper reports quantitative comparisons across datasets and tasks on large test sets with fixed evaluation protocols, and reports standard deviation for certain experiments. Yet, to keep the computation budget at a reasonable level, for a large set of experiments, the current manuscript does not include error bars, confidence intervals, standard deviations, or statistical significance tests. 

\item {\bf Experiments compute resources}
    \item[] Question: For each experiment, does the paper provide sufficient information on the computer resources (type of compute workers, memory, time of execution) needed to reproduce the experiments?
    \item[] Answer: \answerYes{}
    \item[] Justification: The appendix reports that HaNDF training uses a single NVIDIA A100 GPU and takes approximately 8 hours. 
    
\item {\bf Code of ethics}
    \item[] Question: Does the research conducted in the paper conform, in every respect, with the NeurIPS Code of Ethics \url{https://neurips.cc/public/EthicsGuidelines}?
    \item[] Answer: \answerYes{}
    \item[] Justification: The work develops a hand-object interaction prior and evaluates it on established research datasets. The paper does not involve collection of new private data, deceptive practices, or deployment on human subjects.

\item {\bf Broader impacts}
    \item[] Question: Does the paper discuss both potential positive societal impacts and negative societal impacts of the work performed?
    \item[] Answer: \answerYes{} %
    \item[] Justification: HaNDF can benefit applications that require reliable hand--object interaction understanding, including AR/VR interfaces, assistive robotics, dexterous manipulation, and 3D perception under occlusion. By encouraging physically plausible hand poses conditioned on object geometry, the method may reduce reconstruction artifacts such as implausible joint configurations, missing contact, or hand--object penetration. Potential negative impacts arise if accurate hand--object reconstruction is used for privacy-sensitive behavioral tracking or surveillance without consent. The model may also inherit biases from existing HOI datasets, including limited coverage of hand shapes, grasp styles, object categories, and camera viewpoints. Finally, although HaNDF improves plausibility, it is not a safety guarantee; deployment in robotic manipulation or other safety-critical settings would require additional validation, uncertainty estimation, and fail-safe mechanisms.
    
\item {\bf Safeguards}
    \item[] Question: Does the paper describe safeguards that have been put in place for responsible release of data or models that have a high risk for misuse (e.g., pre-trained language models, image generators, or scraped datasets)?
    \item[] Answer: \answerNA{}
    \item[] Justification: The method is a geometric prior for hand-object interaction and does not introduce a high-risk generative model, sensitive dataset, or deployed decision-making system. The paper should nevertheless clarify that the method is intended for research use and should not be deployed in safety-critical robotic manipulation without further validation.

\item {\bf Licenses for existing assets}
    \item[] Question: Are the creators or original owners of assets (e.g., code, data, models), used in the paper, properly credited and are the license and terms of use explicitly mentioned and properly respected?
    \item[] Answer: \answerYes{}
    \item[] Justification: The paper cites the three existing datasets used in the experiments, HOI4D, DexYCB, and OakInk, and includes an appendix paragraph documenting their licenses and non-commercial research use. It also documents the MANO license and the licenses of the relevant dataset toolkits, and states that restricted MANO assets are not redistributed.

\item {\bf New assets}
    \item[] Question: Are new assets introduced in the paper well documented and is the documentation provided alongside the assets?
    \item[] Answer: \answerYes{}
    \item[] Justification: The paper introduces a well documented new method, model checkpoint, and code package as part of the current submission. We provide relevant documentation, license information, and usage instructions alongside the assets.

\item {\bf Crowdsourcing and research with human subjects}
    \item[] Question: For crowdsourcing experiments and research with human subjects, does the paper include the full text of instructions given to participants and screenshots, if applicable, as well as details about compensation (if any)? 
    \item[] Answer: \answerNA{}
    \item[] Justification: The paper does not conduct new crowdsourcing experiments or new human-subject data collection. It evaluates on existing public hand-object interaction datasets.

\item {\bf Institutional review board (IRB) approvals or equivalent for research with human subjects}
    \item[] Question: Does the paper describe potential risks incurred by study participants, whether such risks were disclosed to the subjects, and whether Institutional Review Board (IRB) approvals (or an equivalent approval/review based on the requirements of your country or institution) were obtained?
    \item[] Answer: \answerNA{}
    \item[] Justification: The paper does not involve new human-subject experiments or newly collected participant data. It uses previously released datasets, so IRB information should be inherited from the original dataset papers where applicable.

\item {\bf Declaration of LLM usage}
    \item[] Question: Does the paper describe the usage of LLMs if it is an important, original, or non-standard component of the core methods in this research? Note that if the LLM is used only for writing, editing, or formatting purposes and does \emph{not} impact the core methodology, scientific rigor, or originality of the research, declaration is not required.
    \item[] Answer: \answerNA{} %
    \item[] Justification: The core method development, experiments, and analysis in this work do not use LLMs as an important, original, or non-standard component. Any LLM use, if any, was limited to writing, editing, or formatting assistance and did not affect the scientific methodology, results, or claims.

\end{enumerate}